%% file: paper.tex
\documentclass{article}

\usepackage[main,final]{neurips_2026}
\usepackage[utf8]{inputenc}
\usepackage[T1]{fontenc}
\usepackage{amsmath,amsfonts,amsthm}
\usepackage{booktabs}
\usepackage{graphicx}
\usepackage[table]{xcolor}
\usepackage{microtype}
\usepackage{longtable,needspace}
\usepackage{url}
\usepackage[hidelinks]{hyperref}

\graphicspath{{figures/}}

\newtheorem{proposition}{Proposition}[section]

\definecolor{sharingRed}{HTML}{F8E4DF}
\definecolor{sharingBlue}{HTML}{E5EDF8}
\newcommand{\sharebelow}[1]{\cellcolor{sharingRed}\textbf{#1}}
\newcommand{\shareabove}[1]{\cellcolor{sharingBlue}#1}

\title{Massive Activations Confound Sparse Autoencoders}
\author{%
  Nathan Delisle \\
  University of Chicago \\
  \texttt{ndelisle@uchicago.edu}
  \And
  Chenhao Tan \\
  University of Chicago \\
  \texttt{chenhao@uchicago.edu}
}

\begin{document}
\raggedbottom
\maketitle

\makeatletter
\begingroup
\renewcommand{\thefootnote}{}
\long\def\@makefntext#1{\noindent#1}
\footnotetext{%
Code: \url{https://github.com/nathandelisle/massive-saes}.\\
SAEs: \url{https://huggingface.co/nathandelisle/massive-saes}.
}
\endgroup
\makeatother

\begin{abstract}
Sparse autoencoders (SAEs) are commonly evaluated by variance explained and ablation damage. We show how massive activations undermine both evaluations, and propose a correction. Massive activations cause up to 99.9\% of activation variance to lie along the first principal component (PC1). Because the PC1 component has no unique sparse decomposition, each SAE run rebuilds it from a different set of features. These features dominate importance rankings: an average of 47 of the 50 most important features are unmatched in a separate run, compared with about 27 of a random 50. A dictionary can also report 99.96\% of variance explained while leaving 33\% of the variance outside PC1 unexplained. We \emph{bypass} PC1 by training the SAE on activations with PC1 removed, then adding PC1 to the reconstruction. In affected models, the top-50 feature sharing rate rises from 4--39.5\% to 73--99.5\%. At the same TopK sparsity, the resulting SAE degrades the model up to 44\% less and reduces content reconstruction error by up to 33.9\%.
\end{abstract}

\section{Introduction}
\label{sec:intro}
\begin{figure}[t!]
  \centering
  \begin{minipage}[t]{0.455\textwidth}\centering
        \includegraphics[width=\linewidth]{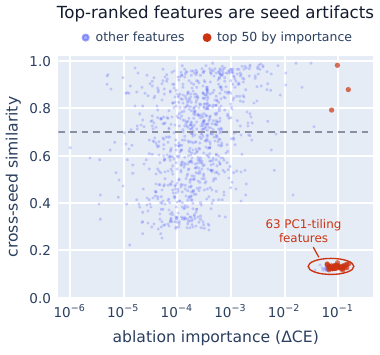}
  \end{minipage}\hfill
  \begin{minipage}[t]{0.545\textwidth}\centering
        \includegraphics[width=\linewidth]{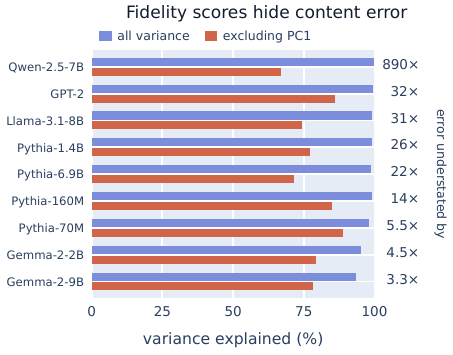}
  \end{minipage}
  \caption{Massive activations confound both standard SAE evaluations. \textbf{Left:} The 50 most important features (red) are seed artifacts: 47 of 50 fail to recur across training runs (GPT-2; dashed line indicates sharing threshold). \textbf{Right:} Variance explained, reported vs.\ excluding PC1. Models are ordered by PC1 dominance. The reported score understates content error by up to $890\times$.}
  \label{fig:hero}
\end{figure}

Sparse autoencoders (SAEs) are trained to reconstruct a language model's internal activations using a sparse set of learned directions, with the aim that each direction corresponds to a meaningful feature used by the model~\citep{cunningham2023sparse, bricken2023monosemanticity}. SAEs are primarily evaluated by variance explained, which measures how well a dictionary reconstructs the activations, and ablation damage, which measures how much the model depends on a single feature~\citep{gao2024scaling, templeton2024scaling}. On these metrics, SAEs appear to succeed: in GPT-2, SAEs explain 99.6\% of activation variance, and removing just one learned feature can raise perplexity by 17\% (Figure~\ref{fig:hero}). However, SAE features are known to be unstable across training seeds: \citet{paulo2025sae} found that about 70\% of features differ across training seeds on the same model and data.

Naively, features that are important to the model should recur more reliably across training seeds, because their directions should be more strongly determined by the model and its training data. We find the opposite: \textbf{the most important features are the least reproducible} (Figure~\ref{fig:hero}, left). In five highly anisotropic models, the 50 most important features have mean pairwise sharing rates of only 5--11\% across training seeds, compared with dictionary-wide sharing rates of 47--63\%.

We trace this inversion to massive activations. In middle layers, 70--99.9\% of activation variance lies along the first principal component (PC1) of the residual stream. This is a known effect of massive activations, outlier dimensions written by early feed-forward layers to implement attention sinks \citep{sun2024massive, xiao2023streaming, sink_valleys}. While PC1 itself is well-defined, under the TopK objective, there are infinitely many ways to exactly reconstruct an isolated rank-one signal (Appendix~\ref{app:theory}). SAEs trained with different seeds therefore ``tile'' PC1 with different sets of features. The tiles' components outside PC1 can cancel, so ablating one tile can cause damage by breaking that cancellation. The tiling can be directly observed by grouping features by their decoder's alignment with PC1. Cross-seed sharing is at the dictionary baseline for unaligned features, is near zero for moderately PC1-aligned features (tiles), and reaches one for the rare ``canonical features'' highly aligned with PC1 (Section~\ref{sec:confound}).

Massive activations also confound variance explained, which is reported equivalently as the fraction of variance unexplained (FVU). Because reconstruction error is variance-weighted, the reported score is dominated by how well the SAE reconstructs PC1. When reconstruction error along PC1 is negligible, the error on the variance outside PC1, or ``content variance,'' is understated by the dominance factor: $\mathrm{FVU}_{\mathrm{content}}/\mathrm{FVU}_{\mathrm{total}} \approx 1/(1-w_1)$, where $w_1$ is PC1's share of activation variance. We measure this approximation and find that it holds on all measured models to within 0.2\% (Section~\ref{sec:fvu}). The reported score can therefore hide substantial content error. Our Qwen-2.5-7B SAEs explain 99.96\% of total variance while leaving 33.0\% of the content variance unexplained (Figure~\ref{fig:hero}, right).

We test how the sparsity objective contributes to this instability. Because TopK constrains only \emph{which} features fire, different decompositions of PC1 achieve the same reconstruction loss within the sparsity budget. Adding a magnitude penalty $\beta\lVert\mathbf z\rVert_1$ raises top-50 sharing from 6.4\% to 94.9\% with less than 1\% change in sparsity or reconstruction error (Section~\ref{sec:cause}). In our experiments, TopK, BatchTopK \citep{bussmann2024batchtopk}, and JumpReLU with scheduled thresholds \citep{rajamanoharan2024jump} show the inversion between reproducibility and importance, whereas ReLU+$\ell_1$ does not. Released dictionaries also show the confound: FVU outside PC1 is approximately $2.1\times$ total FVU in Gemma Scope 2, although normalization substantially reduces this effect in other dictionaries (Section~\ref{sec:released}).

To remove the confound without altering the SAE's sparsity objective, we train on activations with the PC1 projection removed and add the projection back to the SAE's output at inference. With PC1 absent from the SAE's input, top-50 feature sharing rises to 73--99.5\%, and importance predicts reproducibility: the ten most important features share at 1.4--2.0 times the dictionary-wide rate in every model. A post-hoc variant that excludes PC1-aligned features from the ranking reaches the same result without retraining. Where the direction dominates, the resulting SAE is also better outright, with up to 44\% less degradation of the model it is inserted into and 33.9\% less content reconstruction error at the same TopK sparsity (Section~\ref{sec:repair}; Appendix~\ref{app:splice}).

Finally, we establish the scope of the confound. In Pythia-1.4B, the inversion disappears at the final layer, where PC1 accounts for much less activation variance (Section~\ref{sec:depth}). Excluding the first position removes PC1 dominance where massive activations are mainly present there, as in Llama-3.1-8B (97\% to 3\%), while Qwen-2.5-7B remains at 99.7\% (Appendix~\ref{app:bos}). Furthermore, while separating PC1 changes \emph{which} features the metrics select, the dictionary-wide sharing rate remains between 44\% and 68\%.

\section{Background}
\label{sec:background}

\paragraph{Feature reproducibility.}
Because SAE training is non-convex, different random initializations can result in different learned features even when the training data and hyperparameters are fixed. \citet{paulo2025sae} compare independently initialized SAEs trained on identical data and find that only 30\% of features are shared between runs on Llama-3-8B. They also find that at similar sparsity, ReLU+$\ell_1$ SAEs share more features across runs than TopK SAEs do. \citet{gerasimov2026unstable} find that while individual features often fail to recur across runs, the subspaces they span largely do. They find that stable features carry most of the reconstruction- and prediction-relevant signal. \citet{sun2025dense} identify pairs of dense latents with opposing decoder directions aligned with PC1 in Gemma Scope. We identify massive activations as a particularly impactful source of this instability: independently trained SAEs reconstruct the same dominant PC1 component using different features, which can dominate the importance ranking.

\paragraph{SAE evaluation.}
Alongside reconstruction quality, SAEBench evaluates SAE performance using concept detection, feature disentanglement, and automated interpretability, and finds that improvements in the sparsity--fidelity trade-off do not reliably improve these measures \citep{karvonen2025saebench}. \citet{saraswatula2025whitening} find that PCA whitening improves sparse probing for both ReLU and TopK SAEs, despite reducing reconstruction fidelity. \citet{chanin2026benchmarks} identifies evaluation noise and failures to track ground-truth feature recovery in several benchmark metrics. \citet{engels2025darkmatter} show that much of SAE reconstruction error is linearly predictable from the input activations. \citet{timkey2021rogue} show that a few rogue dimensions dominate cosine similarity between contextual embeddings. Our results demonstrate that PC1 plays an analogous role in SAE fidelity.

\paragraph{Massive activations and attention sinks.}
Massive activations are unusually large values concentrated in a few activation dimensions and token positions. \citet{sun2024massive} show that these values can be mostly input-independent and attract attention to the associated tokens, connecting them to attention sinks \citep{xiao2023streaming}. \citet{yu2024superweight} identify individual weights in early MLPs that generate massive activations and whose removal can severely degrade language-model performance. These outliers can cause PC1 to account for most of the activation variance. For uncentered activations, \citet{sink_valleys} show how large token norms can make a single direction account for most of the total squared activation magnitude. In vision transformers, \citet{darcet2024vision} find that high-norm background patches are reused for internal computation, and introduce register tokens so that the model can store intermediate results without using background patch tokens. \citet{jiang2025vision} find the neurons that produce these high-norm patches and move their output into register tokens added at inference, without retraining the model. These studies explain the function and origin of massive activations; we investigate how massive activations affect SAE learning and evaluation.

\section{Experimental Setup}
\label{sec:setup}

\paragraph{SAEs.} We train SAEs on residual stream activations $\mathbf x \in \mathbb R^{d}$ from GPT-2, Pythia-70M-deduped/160M/1.4B/6.9B, Qwen-2.5-7B, Llama-3.1-8B/70B, and Gemma-2-2B/9B. Unless stated otherwise, activations are taken at the output of the middle transformer block over 5M tokens of WikiText-103 \citep{merity2017pointer}, with one document per sequence and padding excluded. We train TopK SAEs \citep{gao2024scaling} with $m=4d$ dictionary features and sparsity $k=64$, using Adam for 20{,}000 steps with a learning rate of $3\times10^{-4}$ and 4096 activations per batch, and renormalizing the decoder directions to unit norm after each step. Initialization is varied across seeds. The appendix lists mini-batch order, seed counts and further information on the training settings.

\paragraph{Metrics.}
Reconstruction fidelity is measured by the fraction of variance unexplained,
$\mathrm{FVU}=\mathbb E\lVert\mathbf x-\hat{\mathbf x}\rVert^2/\mathbb E\lVert\mathbf x-\bar{\mathbf x}\rVert^2$, where $\bar{\mathbf x}$ is the training mean. Let $\mathbf v_1$ be the unit PC1 direction of the centered training activations,
$w_1$ its share of variance in the evaluated population, and
$P_\perp=I-\mathbf v_1\mathbf v_1^\top$.
Content FVU measures reconstruction outside PC1:
$\mathrm{FVU}_{\mathrm{content}}=\mathbb E\lVert P_\perp(\mathbf x-\hat{\mathbf x})\rVert^2/\mathbb E\lVert P_\perp(\mathbf x-\bar{\mathbf x})\rVert^2$.
Feature importance, $I_j$, is the increase in cross-entropy when feature $j$
is set to zero in the SAE reconstruction (relative to the same model with
the non-ablated SAE). Splice cost is the increase in cross-entropy from
replacing the original activation with its SAE reconstruction.

\paragraph{Reproducibility and PC1 alignment.} We measure cross-seed reproducibility using decoder similarity \citep{paulo2025sae}. Feature $j$ is \emph{shared} with another seeded SAE if $\max_l\cos(\mathbf d_j,\mathbf d'_l)\geq0.7$, where the maximum is over all decoder directions in that SAE. The sharing rate is the fraction of selected features satisfying this criterion. Dictionary-wide sharing uses all alive features. A feature's PC1 alignment is $a_j=\cos^2(\mathbf d_j,\mathbf v_1)$.

\begin{figure}[!b]
  \centering
  \includegraphics[width=\linewidth]{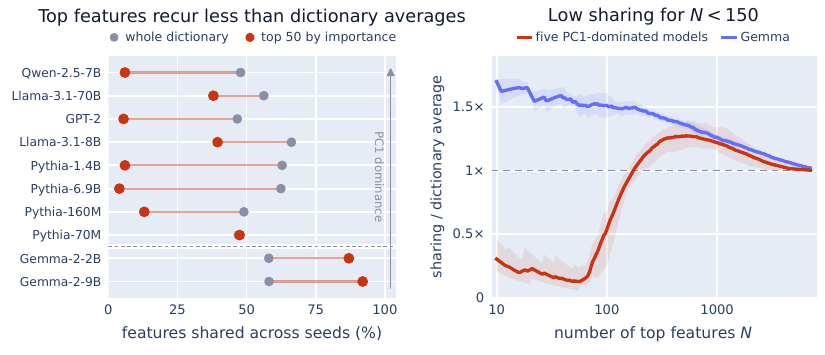}
\caption{Importance selects less reproducible features in seven of ten models. \textbf{Left:} mean sharing across seed comparisons for the top 50 features by importance (red) and the whole dictionary (gray), ordered by PC1 dominance. The Gemma models below the dotted line show the opposite pattern. \textbf{Right:} mean and range of top-$N$ sharing relative to dictionary-wide sharing: GPT-2, Qwen-2.5-7B, and Pythia-160M/1.4B/6.9B (red); Gemma-2-2B/9B (blue). The dashed line denotes equal sharing.}
  \label{fig:inversion}
\end{figure}

\section{The Confound}
\label{sec:confound}

\subsection{Reproducibility of Important Features} \label{sec:inversion} 
For each model, we rank SAE features by zero-ablation importance and compare the mean pairwise sharing rate of the top 50 with that of the full dictionary (Figure~\ref{fig:inversion}, left). In every model where PC1 explains at least 93\% of activation variance ($w_1 \ge 0.93$), the top 50 most important features are shared at a lower rate than the full dictionary. In five of these models (GPT-2, Qwen-2.5-7B, and Pythia-160M/1.4B/6.9B), an average of 43.5--48 of the top 50 features fail to recur in another run, where choosing 50 features at random from the full dictionary would leave an average of 19--27 without a match. Across all models we evaluate, the top-50 sharing rate falls strongly with PC1 dominance (Pearson $r=-0.93$, Spearman $\rho=-0.73$ against $w_1$).

This failure mainly affects the very top of the importance ranking. Across all features, importance and reproducibility are weakly correlated: the rank correlation between importance and reproducibility is between $+0.11$ and $+0.19$ in every model. Although the overall correlation is weak, the top of the ranking is commonly used to select features. For example, ablation-based importance is used to select features when tracing feature circuits \citep{cunningham2023sparse}, and identifying spurious features in classifiers \citep{marks2024sparse}.

It is reasonable to ask whether these results depend on measuring sharing among the top 50 features specifically. Figure~\ref{fig:inversion} (right) varies the number of top features $N$ from 10 to the full dictionary. For the five PC1-dominated models, the mean stays below the dictionary-wide baseline for $10\leq N<150$. In the Gemma models, which do not show the inversion, the ranking already selects ordinary features, so sharing declines as more features are included. Both curves reach 1 at the right edge by construction, where the top $N$ is the whole dictionary. Per-model values are given in Table~\ref{tab:app-topn}.

In a separate replication using GPT-2 and 10 seeds, we find that the confound persists under both cross-entropy and KL rankings. One-to-one matching changes the baseline top-50 sharing rates by less than 0.3 percentage points. Jointly resampling seeds and evaluation articles also does not affect the result. Our controls are described in Appendix~\ref{app:native}.

Gemma-2-2B and Gemma-2-9B, the two models with the lowest PC1 dominance ($w_1=0.78$, $0.70$), have top-50 sharing rates of 87\% and 92\%. This result is consistent with the confound depending on PC1. Llama-3.1-70B and Llama-3.1-8B have top-50 sharing rates of 38\% and 56\%, below their dictionary-wide rates but above the fully inverted models. In these models, a smaller fraction of ablation importance falls on PC1-aligned features. This is because Llama's outlier signal is concentrated at the BOS position \citep{sun2024massive}, so its PC1-tiling features fire almost exclusively there (62 features, 3\% shared across seeds; Appendix~\ref{app:bos}). Qwen shows a stronger inversion, with top-50 sharing of 6\%.

\subsection{Reproducibility and PC1 Alignment}
\label{sec:tiles}

\begin{figure}[b!]
  \centering
  \includegraphics[width=\linewidth]{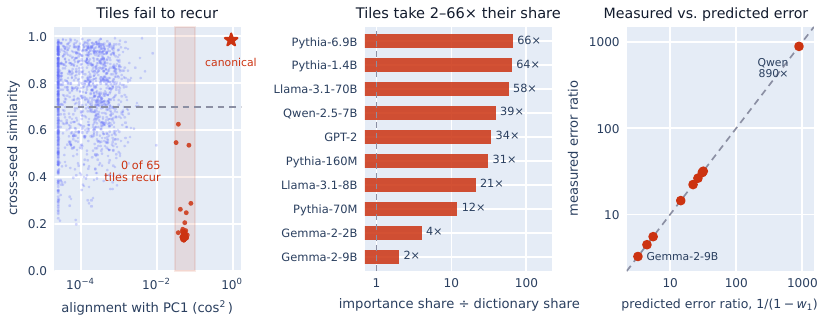}
  \caption{\textbf{Left:} cross-seed similarity (best-match decoder cosine) against PC1 alignment for every sampled feature in Pythia-1.4B. 
  \textbf{Middle:} the tiles' share of total ablation importance divided by their share of the dictionary, for all ten models. \textbf{Right:} the measured ratio $\mathrm{FVU}_{\mathrm{content}}/\mathrm{FVU}_{\mathrm{total}}$ against the prediction $1/(1-w_1)$ for standard SAEs across nine models, using the same held-out evaluation as Figure~\ref{fig:hero}.}
  \label{fig:confound}
\end{figure}

We group each model's features by their decoder's alignment with PC1 and measure the sharing rate within each group (Figure~\ref{fig:confound}, left). The resulting features fall into three categories. Features with alignment $a_j=\cos^2(\mathbf d_j,\mathbf v_1)$ below $10^{-2}$ share at the normal, dictionary-wide rate. Features at intermediate alignment share at 0.0--2.6\%: these are the features that jointly reconstruct PC1, which we call tiles. In every model, the tiles occupy one contiguous range of alignment, beginning at 0.01 in GPT-2, Qwen-2.5-7B, and Llama-3.1-70B and at 0.03 in the other seven models. Each model has 38--106 tiles, about 1\% of all alive features (Appendix~\ref{app:composition}).

The third category is the canonical feature: a feature that is highly-aligned with PC1 (alignment at least 0.6). Four of the ten dictionaries contain exactly one, Gemma-2-2B contains two, and the remaining five contain none. Where present, canonical features recur in every run.

Tiles make up about 1\% of alive features, but are responsible for 2--66 times their share of total ablation importance, with the two lowest values in the Gemma family (Figure~\ref{fig:confound}, middle). In the most affected models, these irreproducible features occupy the top of the importance ranking, explaining the inversion in Section~\ref{sec:inversion}.

We test whether these large ablation effects arise from removing PC1 by separately ablating the PC1 and orthogonal components of each top-50 feature's reconstruction contribution. Removing only the orthogonal component preserves the SAE's PC1 reconstruction but reproduces 88--100\% of the mean whole-feature cross-entropy increase across GPT-2, Pythia-160M/1.4B, and Qwen-2.5-7B (Appendix~\ref{app:component_ablation}), consistent with ablation disrupting cancellation between features outside PC1.

\subsection{Fidelity and the Dominance Factor}
\label{sec:fvu}
Because FVU is a variance-weighted average over directions, it can be split along PC1. Writing $\mathrm{FVU}_{\mathrm{PC1}}$ for the same ratio computed on the PC1 component alone, $\mathrm{FVU}_{\mathrm{total}} = w_1\,\mathrm{FVU}_{\mathrm{PC1}} + (1-w_1)\,\mathrm{FVU}_{\mathrm{content}}$. The SAE reconstructs PC1 nearly exactly: error along PC1 accounts for less than 0.3\% of total reconstruction error. Neglecting this contribution, the reported score understates content error by approximately the dominance factor: $\mathrm{FVU}_{\mathrm{content}}/\mathrm{FVU}_{\mathrm{total}} \approx 1/(1-w_1)$. 

This approximation has a relative error below 0.2\% (Figure~\ref{fig:confound}, right; full values in Appendix~\ref{app:fvutable}). The reported score therefore conceals substantial reconstruction error outside PC1. Our Qwen-2.5-7B SAEs explain 99.96\% of activation variance while leaving 33.0\% of the content variance unexplained, an understatement of approximately $890\times$. Figure~\ref{fig:hero} (right) shows this comparison across all models.

Prior work has argued that some model structure resists sparse linear decomposition \citep{engels2024notall, bhalla2026manifolds}. In our experiments, while SAEs reconstruct content better than PCA baselines, they still leave substantial error. At 64 active coefficients, our standard SAEs outperform both rank-64 PCA and the adaptive 64-component PCA baseline, but still leave 12--33\% of the content variance unexplained (Appendix~\ref{app:baselines}). While bypass improves this reconstruction, reducing content error by 33.9\% in Qwen-2.5-7B and by 0.8--4.9\% in the other tested models (Section~\ref{sec:repair}), a substantial fraction of the content variance remains unexplained.

\section{Cause of the Confound}
\label{sec:cause}
The PC1 component can be reconstructed exactly by one aligned feature or several partially aligned features whose components outside PC1 cancel. For an isolated PC1 component, either representation achieves zero reconstruction loss within the sparsity budget; TopK adds no penalty on activation magnitudes.

We test this explanation by adding an activation-magnitude penalty $\beta\lVert\mathbf z\rVert_1$ to the loss. Because our decoder directions have unit norm, a partially aligned feature needs a larger activation to contribute the same amount to PC1. For an exact reconstruction of PC1, partially aligned features require a larger sum of activation magnitudes, so the $\ell_1$ penalty prefers PC1-aligned features (Appendix~\ref{app:theory}). We test whether this change makes important features more reproducible, then compare the pattern across SAE objectives.

\begin{figure}[b!]
  \centering
  \begin{minipage}[t]{0.47\textwidth}\centering
    \includegraphics[width=\linewidth]{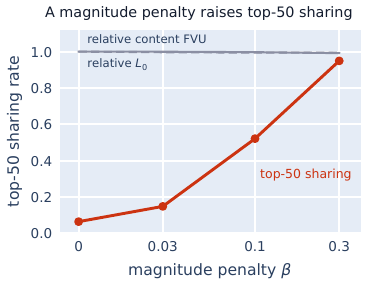}
  \end{minipage}\hfill
  \begin{minipage}[t]{0.47\textwidth}\centering
    \includegraphics[width=\linewidth]{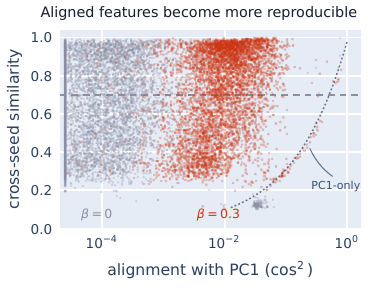}
  \end{minipage}
  \caption{\textbf{Left:} top-50 sharing on GPT-2 against the magnitude penalty $\beta$. Sharing rises from 0.06 to 0.95, while $L_0$ and content FVU stay within 1\%; both controls normalized to $\beta=0$. \textbf{Right:} cross-seed similarity against PC1 alignment for every feature at $\beta=0$ (gray) and $\beta=0.3$ (red), one point per feature. The dashed line indicates the sharing threshold. The dotted curve is the similarity two features obtain from their PC1 components alone.}
  \label{fig:cause}
\end{figure}

\subsection{Objective and Canonical Initialization}
\label{sec:beta}
We add a penalty $\beta\lVert\mathbf z\rVert_1$ to the TopK objective and measure the reproducibility of important features while tracking sparsity and reconstruction loss. On GPT-2 at block 6, we train five seeds at four levels of magnitude penalty $\beta \in \{0, 0.03, 0.1, 0.3\}$, keeping the remaining training settings fixed. At $\beta=0.3$, mean pairwise sharing among the top 50 features rises from 6.4\% to 94.9\%. Sparsity and reconstruction stay roughly constant: mean $L_0$ goes from 64.00 to 63.47, and content FVU goes from 0.1394 to 0.1387 (Figure~\ref{fig:cause}, left).

Features with PC1 alignment $0.01\leq a_j<0.3$ still make up 73\% of the top 50 at $\beta=0.3$, compared to 93.6\% of the top 50 without the penalty. Their number increases from 65 to 1{,}307 per seed, but their sharing rate also rises from 3\% to 72\%. Full results are in Appendix~\ref{app:beta}.

The magnitude penalty also increases cross-seed similarity in the decoder components outside PC1. For a fixed pair of seeds, we match the top-50 features by decoder cosine and then project PC1 out of both matched directions. Mean cosine similarity between the remaining components rises from 0.165 to 0.831 on GPT-2 as $\beta$ increases from 0 to 0.3, and from 0.114 to 0.910 on Pythia-1.4B as $\beta$ increases from 0 to 0.1 (Appendix~\ref{app:beta}).

We also test whether initializing the SAE with a single PC1-aligned feature prevents PC1 tiling under the original TopK objective. We initialize one feature's decoder direction along PC1 and do not hold it fixed during training. In a GPT-2 experiment with 10 seeds, the top-50 sharing rate rises from 5.5\% to 68\%, and the median tile count drops from 65 to one. Canonical initialization worsens reconstruction loss, however: median content FVU rises from 0.149 to 0.165. Doubling dictionary width with standard initialization leaves the median tile count at 64 (Appendix~\ref{app:native}).

\subsection{Reproducibility Across SAE Objectives}
\label{sec:objective}

\begin{table}[!t]
  \setlength{\abovecaptionskip}{7pt}
  \setlength{\belowcaptionskip}{0pt}
  \centering
  \small
  \setlength{\tabcolsep}{4pt}
  \renewcommand{\arraystretch}{1.15}
  \begin{tabular*}{\textwidth}{@{\extracolsep{\fill}}lrrrrrr@{}}
    \toprule
    Model & $w_1$ & TopK & BatchTopK & JumpReLU$^{*}$ & Gated & ReLU+$\ell_1$ \\
    \midrule
    Qwen-2.5-7B & .999 & \sharebelow{6.0} & \sharebelow{33.0} & \sharebelow{32.0} & \shareabove{47.0$^\dagger$} & \shareabove{97.0} \\
    GPT-2 & .970 & \sharebelow{5.5} & \sharebelow{17.5} & \sharebelow{11.5} & \sharebelow{11.0} & \shareabove{100.0} \\
    Llama-3.1-8B & .969 & \sharebelow{56.0} & \sharebelow{62.0} & \sharebelow{61.0} & \shareabove{99.0} & \shareabove{97.0} \\
    Pythia-1.4B & .964 & \sharebelow{10.0} & \sharebelow{14.0} & \sharebelow{12.0} & \shareabove{86.0} & \shareabove{98.5} \\
    Pythia-6.9B & .957 & \sharebelow{10.0} & \sharebelow{18.0} & \sharebelow{26.0} & \sharebelow{15.0$^\ddagger$} & \shareabove{97.0} \\
    Pythia-160M & .934 & \sharebelow{9.0} & \sharebelow{9.5} & \sharebelow{14.5} & \shareabove{87.0} & \shareabove{96.5} \\
    \midrule
    Gemma-2-2B & .781 & \shareabove{80.5} & \shareabove{72.5} & \shareabove{76.0} & \shareabove{79.0} & \shareabove{97.0} \\
    Gemma-2-9B & .700 & \shareabove{90.0} & \shareabove{90.0} & \shareabove{75.0} & \shareabove{63.0} & \shareabove{95.0} \\
    \bottomrule
  \end{tabular*}
  \caption{Top-50 sharing (\%), averaged over comparisons between a
  reference seed and the other training seeds. Red, bold entries fall
  below the corresponding dictionary-wide sharing rate; blue entries
  exceed or match it.
  $^{*}$Scheduled thresholds.
  $^\dagger$One Qwen Gated seed diverged.
  $^\ddagger$Sensitive to sparsity calibration
  (Appendix~\ref{app:archgrid}).}
  \label{tab:objectives}
\end{table}

Table~\ref{tab:objectives} shows the sharing rates of the top-50 most important features across different SAE objectives. For TopK, BatchTopK, and JumpReLU with scheduled thresholds, in models where PC1 explains at least 93\% of variance, the top 50 features are shared 7--53 percentage points below the dictionary-wide sharing rate (Table~\ref{tab:objectives}). ReLU+$\ell_1$, which directly penalizes the reconstruction coefficients, has top-50 sharing of 95--100\% in every model, consistently above its dictionary-wide rate. Neither Gemma model shows the inversion under any objective.

Gated's sparsity penalty acts on gate activations, giving no gradient to the magnitude-only parameters \citep{rajamanoharan2024gated}. We test whether this contributes to seed dependence by adding a penalty on the reconstruction coefficients. On GPT-2, adding a penalty $\beta=0.3$ raises top-50 sharing from 11\% to 98\%. Mean held-out content FVU decreases from 0.157 to 0.146, while mean $L_0$ increases from 56.8 to 59.9. At $\beta=0.1$, we apply the added penalty's gradient only to the magnitude parameters or only through the shared encoder. Top-50 sharing is 89\% and 10\%, respectively (Appendix~\ref{app:archgrid}).

\section{Repair}
\label{sec:repair}

\subsection{Bypass}
\label{sec:bypass}

We compute a unit PC1 direction $\mathbf v_1$ from centered training activations. For each centered activation $\mathbf x$, we subtract $(\mathbf v_1^\top\mathbf x)\mathbf v_1$ and train the SAE to reconstruct the remaining activation. Before passing the reconstructed activation to the language model, we add the removed PC1 component and training mean back. This preserves the signal that the underlying language model uses for attention sinks, while allowing the SAE to reconstruct the remaining structure.

In all models that show the inversion, top-50 sharing increases, from 4--39.5\% to 73--99.5\% (Figure~\ref{fig:repair}, left). The two Gemma models already have highly reproducible top-ranked features and show smaller changes: 87\% to 93.5\% on Gemma-2-2B and 92\% to 94.5\% on Gemma-2-9B.

Because removing PC1 alters the SAE's training inputs, improved reproducibility could come at the cost of less accurate reconstruction of the remaining activation. We test this by comparing held-out content FVU between standard and PC1-bypass SAEs. Both SAEs use a TopK budget of 64 active features per token. Bypass also carries a separate PC1 coefficient, computed directly from the input.

Content reconstruction improves under bypass. Held-out content FVU decreases by 0.8--33.9\% in every model (Figure~\ref{fig:repair}, right). The gain likely comes from freed SAE capacity: on massive-activation tokens, a standard SAE spends about 63 of its 64 active coefficients on features with $a_j \geq 0.01$ in GPT-2, Pythia-1.4B, and Qwen-2.5-7B. Bypass also improves reconstruction on ordinary tokens, which account for approximately 84\% of GPT-2's reduction in content error. Giving bypass 63 sparse coefficients plus the PC1 coefficient (the same total budget as the standard SAE) retains 68--80\% of the content reconstruction gain on GPT-2, Pythia-1.4B, and Gemma-2-2B, and 99.4\% on Qwen-2.5-7B (Appendix~\ref{app:capacity}).

\paragraph{Resample ablation.} 
Zeroing a feature could itself cause the instability we observe. We rank features by resample ablation, which replaces a feature's activation with draws from its own activation distribution, under marginal resampling (at every token) and support-preserving resampling (nonzero draws only where the feature fires). On GPT-2, marginal resampling results in a top-50 sharing rate of 4.0\% before bypass and 77.5\% after. Support-preserving resampling results in top-50 rates of 83.0\% and 85.0\% respectively. In Pythia-1.4B, the support-preserving rates are 27.5\% and 95.5\%. Thus, instability persists under nonzero replacement, although the extent to which this is true depends on the specific replacement used (Appendix~\ref{app:rx_resampling}).

\subsection{Post-hoc Feature Selection}
\label{sec:posthoc}

PC1 bypass is a substantial intervention: it requires retraining the SAE on activations with PC1 removed, which changes the features it learns. For an existing SAE, a more conservative alternative is to exclude PC1-aligned features from the importance ranking post-hoc.

\begin{figure}[!t]
  \centering
  \begin{minipage}[t]{0.49\textwidth}\centering
    \includegraphics[width=\linewidth]{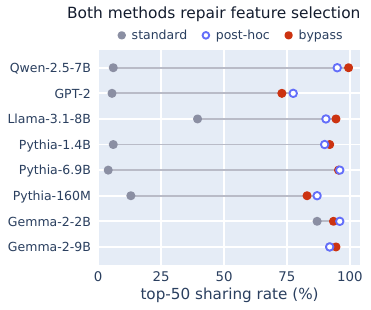}
  \end{minipage}\hfill
  \begin{minipage}[t]{0.49\textwidth}\centering
    \includegraphics[width=\linewidth]{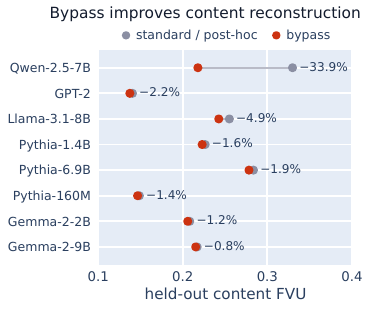}
  \end{minipage}
  \caption{\textbf{Left:} top-50 sharing for standard TopK SAEs, for the same dictionaries with PC1-aligned features excluded from the ranking (post-hoc), and for SAEs trained with PC1 bypass. \textbf{Right:} held-out content FVU for the standard SAE and for bypass, with the relative reduction. Because post-hoc selection uses the standard dictionary, its fidelity is the standard value.}
  \label{fig:repair}
\end{figure}

We exclude features with squared PC1 alignment $a_j\geq0.01$ from the ranking pool, targeting the PC1-tiling features
described in Section~\ref{sec:tiles}. This removes a small population of features (only 0.72--3.92\% of alive features) from consideration, while the 50 most important remaining features are shared at 77.5--96\% across seeds, comparable to the rates obtained with trained bypass (Figure~\ref{fig:repair}, left). Because the dictionary is unchanged, its reconstruction remains identical.

On GPT-2, the selected top 50 achieve 77.5\% cross-seed sharing and have a mean ablation effect of $\Delta\mathrm{CE}=0.0130$. Their individual effects are smaller than those of the original, confounded top 50 ($0.0966$), but similar to those obtained with trained bypass ($0.0141$).

\subsection{Interpretability and Sparse Probing}
\label{sec:interpretability_probing}

\paragraph{Automated interpretability.}
We evaluate the 100 most important features and 100 randomly selected alive features per condition, using one middle-layer SAE seed on each of eight models. An instruction-tuned model explains each feature from activating contexts; a second model scores whether that explanation predicts activation on held-out contexts. We exchange the explainer and evaluator roles between Qwen2.5-32B-Instruct and Mistral-Small-24B-Instruct-2501. For the top-ranked features, mean AUROC against random inactive contexts increases after bypass on GPT-2, all three evaluated Pythia models, and Qwen-2.5-7B in both directions. On Pythia-1.4B, it rises from 0.551 to 0.732 with Qwen explaining and Mistral scoring, and from 0.669 to 0.743 with Mistral explaining and Qwen scoring. Scores fall on Llama-3.1-8B and Gemma-2-2B and rise slightly on Gemma-2-9B (Appendix~\ref{app:rx_autointerp}).

\paragraph{Sparse probing.}
We compare standard and bypass features by fitting logistic probes on SST-2, AG News, Amazon Polarity, and Emotion. With the 20 standard or bypass features of highest training-split mutual information, and mean pooling over up to 64 tokens, the bypass--standard difference in accuracy averaged over the four tasks is less than one percentage point in magnitude for every model. Single model--task differences run from $-2.44$ to $+3.13$ points (Appendix~\ref{app:rx_probing}).

\subsection{Behavioral Interventions}
\label{sec:behavioral}

We test whether the corrected feature importance ranking gives better features for steering, using subject--verb agreement as the task. We greedily select five of each ranking's top 50 features and patch their activations to flip a sentence's subject--verb agreement, with a control that patches from a sentence where only a non-subject noun changed (setup in Appendix~\ref{app:behavioral}). Success requires switching the model's preference between \texttt{is} and \texttt{are} after the subject edit, while leaving it unchanged after the control edit. Averaged over three seeds, post-hoc selection raises success from 0\% to 9.0\%, 31.9\%, and 4.2\% on GPT-2, Pythia-1.4B, and Qwen-2.5-7B; bypass gives success rates of 7.1\%, 37.5\%, and 2.0\%. Gemma-2-2B's success rate is constant at 64--65\% regardless of condition. Features matched across seeds give the same rates to within 0.5 points.

\section{Prevalence}
\label{sec:prevalence}

\subsection{Released SAE Dictionaries}
\label{sec:released}

We compare total and content FVU for publicly released SAE dictionaries \citep{lieberum2024gemmascope,he2024llamascope,mcdougall2025gemmascope2,deng2026qwenscope} to assess how accurately they reconstruct activations outside PC1 (Figure~\ref{fig:prevalence}, left). The median content-to-total FVU ratios are 2.1 for Gemma Scope 2, 1.49 for Pythia, and 98.8 for unfiltered Qwen Scope. Total and content FVU are much closer in Gemma Scope, Llama Scope, and OpenAI's GPT-2 dictionaries, with median ratios of 1.02--1.05.

Preprocessing helps explain these differences. OpenAI's GPT-2 SAEs center and normalize each token's activation before encoding \citep{gao2024scaling}. For the same GPT-2 dictionary, the ratio is 10.9 in the original activation space and 1.03 in the normalized space. \citet{deng2026qwenscope} remove extreme-norm activations when training Qwen Scope, but do not specify the cutoff that they use. We test three cutoffs: 5, 10, and 20 times the median activation norm. Each removes about 0.8\% of activations at the early and middle layers tested. Without these extreme-norm activations, the content-to-total FVU ratios fall from 92--220 to about 1.03.

We train SAEs on Qwen-2.5-7B with regularized PCA whitening and compare them with PC1 bypass at 64 total coefficients. Whitening lowers content error by 39.4\% relative to the standard SAE; bypass lowers content error by 37.6\%. Total error increases to 2.4 times the standard value with whitening but decreases under bypass (Appendix~\ref{app:preproc}).

\subsection{Variation Across Layers}
\label{sec:depth}

In a separate experiment, PC1 explains 93--94\% of activation variance in Pythia-1.4B at layers 3 and 12. Bypass raises top-50 sharing from 14\% to 68\% at block 3 and from 8\% to 92\% at block 12. At final block 23, where PC1 explains only 11\% of variance, top-50 sharing is already 96\% without bypass and does not improve under bypass.

\subsection{Vision Transformers}
\label{sec:vision}

\begin{figure}[!t]
  \centering
  \begin{minipage}[t]{0.47\textwidth}\centering
    \includegraphics[width=\linewidth]{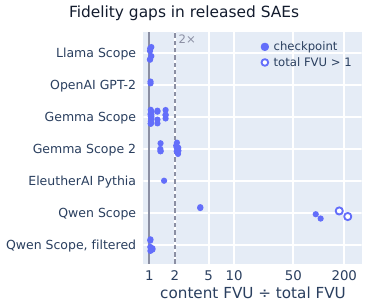}
  \end{minipage}\hfill
  \begin{minipage}[t]{0.47\textwidth}\centering
    \includegraphics[width=\linewidth]{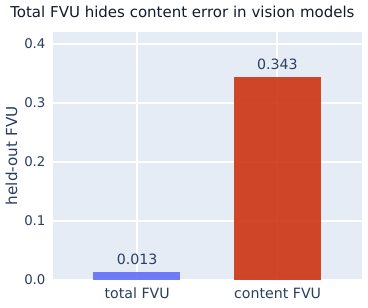}
  \end{minipage}
\caption{\textbf{Left:} content-to-total FVU ratios for released SAEs on held-out text. Each point denotes a single checkpoint. Qwen Scope is shown with and without approximate norm filtering. Hollow markers indicate total FVU above one. Normalization and token-selection conventions are described in Appendix~\ref{app:released}. \textbf{Right:} total and content FVU for BatchTopK SAEs trained on DINOv2-base block-8 patch activations, averaged over three seeds after 10{,}000 training steps.}
  \label{fig:prevalence}
\end{figure}

In DINOv2-base, a self-supervised vision transformer \citep{oquab2023dinov2}, PC1 explains 96.2\% of the variance in patch-token activations. BatchTopK SAEs achieve a total FVU of 0.013, while content FVU is 0.343---approximately 26 times larger (Figure~\ref{fig:prevalence}, right). PC1 bypass modestly improves content FVU to 0.333, compared with 0.344 under random-direction bypass. Pretrained and test-time registers, following \citep{darcet2024vision} and \citep{jiang2025vision}, reduce PC1's share of patch-activation variance to approximately 4\% and nearly eliminate the gap between total and content FVU. Full results are reported in Appendix~\ref{app:vision}.

\paragraph{Crosscoders.} We train crosscoders on four pairs of activation sets from Pythia-1.4B. With PC1 bypassed in both inputs, 83--93\% of the top 50 features are shared across seeds, up from 3--4\%. Normalizing each input or bypassing a random direction does not give comparable gains (Appendix~\ref{app:crosscoders}).

\section{Conclusion}
\label{sec:discussion}

Massive activations confound both feature importance and reconstruction loss in sparse autoencoder evaluation. Accurately reconstructing a dominant PC1 component can hide substantial error in reconstructing the remaining activation, and selecting features with large ablation effects often yields seed-dependent, PC1-reconstructing features. 

PC1 bypass improves content reconstruction and the reproducibility of top-ranked features in affected models. For existing dictionaries, post-hoc selection is a simple way to obtain both reproducible and important features without retraining or altering the existing dictionary. Our results show how accounting for dominant activation structure can improve evaluation and feature selection. More broadly, our findings motivate testing whether other interpretability methods are similarly confounded by dominant structure.   

\begin{ack}
This work was supported in part by compute credits from Amazon Web Services through the AWS Activate program.

The authors declare no competing interests.
\end{ack}

\appendix
\setlength{\LTcapwidth}{\textwidth}
\newenvironment{apptable}[4]{%
  \begingroup\footnotesize\setlength{\tabcolsep}{3pt}%
  \begin{longtable}{@{}#1@{}}%
  \caption{#2}\label{#3}\\
  \toprule #4\\\midrule\endfirsthead
  \toprule #4\\\midrule\endhead
  \bottomrule\endlastfoot
}{\end{longtable}\endgroup}

\section{Experimental Methods}
\label{app:details}

\subsection{Models, activations, and training data}

Table~\ref{tab:app-models} describes where activations are taken from. In all models, transformer layers are numbered starting from 0. Unless output normalization is explicitly stated, we take activations after the layer's attention and MLP, before the next layer's normalization. Pythia-70M uses the deduplicated checkpoint, whereas the larger Pythia models use the standard checkpoints.

\begin{apptable}{lrrrr}{Models, layers used to collect activations, and default SAE widths.}{tab:app-models}{Model & Layers & Layer used & $d$ & $m=4d$}
GPT-2 & 12 & 6 & 768 & 3,072 \\
Pythia-70M & 6 & 3 & 512 & 2,048 \\
Pythia-160M & 12 & 6 & 768 & 3,072 \\
Pythia-1.4B & 24 & 12 & 2,048 & 8,192 \\
Pythia-6.9B & 32 & 16 & 4,096 & 16,384 \\
Llama-3.1-8B & 32 & 16 & 4,096 & 16,384 \\
Llama-3.1-70B & 80 & 40 & 8,192 & 32,768 \\
Qwen-2.5-7B & 28 & 14 & 3,584 & 14,336 \\
Gemma-2-2B & 26 & 13 & 2,304 & 9,216 \\
Gemma-2-9B & 42 & 21 & 3,584 & 14,336 \\
\end{apptable}

By default, we train five standard and five bypass SAEs per model, using seeds $\{42,137,0,1,2\}$. Some experiments use fewer seeds. Seed counts are given alongside the experimental results.

We train SAEs on five million token activations from the \texttt{wikitext-103-v1} training split of WikiText-103 \citep{merity2017pointer}. Each nonempty text entry is tokenized separately, truncated to 128 tokens, and right-padded within batches. Default special tokens are retained; padding activations are excluded. Language models run in evaluation mode in float16, except Gemma, which uses bfloat16. SAEs are trained in float32.

Before SAE training, we subtract the mean $\boldsymbol\mu$ of these five million activations. We estimate PC1, $\mathbf v_1$, by a float64 covariance eigendecomposition of the first 50,000 activations centered by this mean. We use $P=\mathbf v_1\mathbf v_1^\top$ to project onto PC1 and $Q=I-P$ to remove it.

\subsection{SAE training and bypass}

We use TopK SAEs \citep{gao2024scaling} with $m=4d$ features and $k=64$ unless stated otherwise. Each column of the decoder matrix $D$ is a feature direction:
\begin{align}
 \mathbf z(\mathbf x)&=\operatorname{TopK}_{k}\!\left(
 \operatorname{ReLU}\!\left[W_{\rm enc}
 (\mathbf x-\boldsymbol\mu-\mathbf b_{\rm dec})+\mathbf b_{\rm enc}\right]\right),\\
 \widehat{\mathbf x}&=\boldsymbol\mu+\mathbf b_{\rm dec}+D\mathbf z(\mathbf x).
\end{align}
TopK keeps at most $k$ positive entries. We sample Gaussian decoder directions and normalize each to unit length. The encoder is initialized as $D^\top$, but its weights and the decoder weights are updated separately. Biases start at zero.

\begin{apptable}{lp{.55\linewidth}}{Default SAE training settings.}{tab:app-training}{Setting & Value}
Training activations & 5,000,000 \\
Dictionary width & $4d$ \\
TopK budget & 64 \\
Optimizer & Adam \\
Learning rate & $3\times10^{-4}$ \\
Adam moments & $(0.9,0.999)$ \\
Adam epsilon & $10^{-8}$ \\
Steps & 20,000 \\
Batch size & 4,096 \\
Sampling & With replacement \\
Loss & Mean squared reconstruction error \\
Decoder normalization & Unit-length columns after every update \\
\end{apptable}

The same seed sets the initial weights and minibatch order, unless we fix the order across runs. The bypass SAE takes $Q(\mathbf x-\boldsymbol\mu)$ as input. Writing its coefficients as $\mathbf z_B$, we reconstruct
\begin{equation}
 \widehat{\mathbf x}_{B}=\boldsymbol\mu+\mathbf b_{\rm dec}
       +D\mathbf z_{B}+P(\mathbf x-\boldsymbol\mu).
 \label{eq:app-bypass}
\end{equation}
We do not project the decoder output off PC1.

\subsection{Evaluation data}

For feature importance, resampling, PC1 prediction, and automated interpretability, we select WikiText articles outside the SAE training sample using seed 20261005. These articles are divided into non-overlapping sets (Table~\ref{tab:app-partitions}). The token counts are targets based on GPT-2 tokenization; other models use the same articles but may produce different counts. Paragraphs are truncated to 128 tokens. Cross-entropy is averaged over valid next-token predictions, excluding padding and positions without a target.

\begin{apptable}{lrp{.47\linewidth}}{WikiText evaluation sets and target token counts.}{tab:app-partitions}{Article set & Target tokens & Use}
Calibration & 65,536 & Resampling and fitting controls \\
Ranking & 32,768 & Ranking features and component ablations \\
Validation & 32,768 & Validation \\
PC1 prediction test & 65,536 & Testing predictions of model quantities \\
Explanation & 500,000 & Writing feature descriptions \\
Detection & 500,000 & Testing feature descriptions \\
\end{apptable}

\subsection{Feature matching and uncertainty}

To match feature $j$ from seed $s$ against seed $t$, we find its largest decoder cosine:
\begin{equation}
 c_j^{s\to t}=\max_\ell
 \frac{(\mathbf d_j^s)^\top\mathbf d_\ell^t}
 {\|\mathbf d_j^s\|_2\|\mathbf d_\ell^t\|_2}.
 \label{eq:app-sharing}
\end{equation}
A feature is shared if $c_j^{s\to t}\geq0.7$. We do not take the absolute value of cosine. Several source features may match the same target, and we search all target decoder directions unless stated otherwise. Dictionary-wide sharing averages over alive source features; top-$N$ sharing uses the $N$ highest-ranked alive features.

For the default SAEs, the original code counted feature firings over 106,496 training activations but divided by 100,000. We keep the resulting feature selection: the $10^{-3}$ cutoff corresponds to a firing rate above approximately $9.39\times10^{-4}$. We correct mean $L_0$ values from this counter by dividing by 1.06496. Experiments using direct nonzero counts are identified separately.

Unless stated otherwise, sharing is averaged over comparisons from seed 42 to seeds 137, 0, 1, and 2. We also report the average over all 20 comparisons between the five seeds, counting both directions. For confidence intervals, we resample the 75 ranking articles with replacement 1,000 times. Each sample is used to rerank features in both SAEs.

\subsection{Compute resources}
\label{app:compute}

The original experiments ran on Modal H100 GPUs with 128\,GiB of host memory. Llama-3.1-70B used two H200 GPUs and 256\,GiB. Later experiments ran on AWS A10G, L4, L40S and RTX PRO 6000 GPUs, including two RTX PRO 6000s for the 70B evaluation.

Recorded evaluations of linear baselines and model loss with SAE reconstructions took about 2--28 minutes per model. Hardware configurations and available run records are included in the supplement.

\section{Feature Reproducibility and Importance}
\label{app:composition}

\subsection{Zero ablation of all alive features}

We remove one feature at a time from the SAE reconstruction and rank all alive features by the resulting increase in cross-entropy.

\begin{apptable}{lrrrr}{Sharing (\%). S/B denotes standard/bypass. Post-hoc ranks only standard-SAE features with $a_j<.01$.}{tab:app-sharing}{Model & Dictionary S/B & Top 50 S/B & All-pairs top 50 S/B & Post-hoc}
GPT-2 & 46.6/44.3 & 5.5/73.0 & 5.4/72.7 & 77.5 \\
Pythia-70M & 47.0/45.5 & 47.5/81.0 & 43.2/77.9 & 81.5 \\
Pythia-160M & 49.0/46.5 & 13.0/83.0 & 13.3/86.1 & 87.0 \\
Pythia-1.4B & 62.9/58.7 & 6.0/92.0 & 6.8/88.4 & 90.0 \\
Pythia-6.9B & 62.4/61.1 & 4.0/95.5 & 4.4/96.1 & 96.0 \\
Llama-3.1-8B & 66.2/68.0 & 39.5/94.5 & 45.7/93.8 & 90.5 \\
Qwen-2.5-7B & 47.8/61.4 & 6.0/99.5 & 6.0/96.8 & 95.0 \\
Gemma-2-2B & 58.1/62.1 & 87.0/93.5 & 84.1/93.6 & 96.0 \\
Gemma-2-9B & 58.1/60.6 & 92.0/94.5 & 88.3/94.7 & 92.0 \\
\end{apptable}

\begin{apptable}{lrrr}{95\% confidence intervals from resampling articles: top-50 sharing (\%) and the bypass--standard difference (percentage points).}{tab:app-sharing-ci}{Model & Standard & Bypass & Difference}
GPT-2 & [5.5, 5.5] & [69.5, 78.5] & [64.0, 73.0] \\
Pythia-70M & [40.5, 52.5] & [74.0, 83.0] & [25.0, 39.5] \\
Pythia-160M & [9.0, 13.0] & [80.5, 86.5] & [68.0, 75.5] \\
Pythia-1.4B & [4.0, 8.0] & [88.5, 93.5] & [82.0, 88.0] \\
Pythia-6.9B & [4.0, 6.0] & [92.0, 99.0] & [87.0, 95.0] \\
Llama-3.1-8B & [36.5, 46.5] & [91.0, 96.5] & [47.0, 58.5] \\
Qwen-2.5-7B & [6.0, 6.0] & [98.5, 99.5] & [92.5, 93.5] \\
Gemma-2-2B & [81.0, 90.5] & [88.5, 95.5] & [1.5, 11.0] \\
Gemma-2-9B & [90.0, 96.5] & [92.0, 94.5] & [-3.0, 4.0] \\
\end{apptable}

For Llama-3.1-70B, we use seeds 42 and 137. Standard dictionary-wide sharing is 56\%; top-50 sharing rises from 38\% to 84\% with bypass.

\subsection{Post-hoc selection and top-$N$ sharing}
\label{app:posthoc}

Post-hoc selection removes features with $a_j\geq.01$ from the ranking pool, leaving the SAE itself unchanged.

\begin{apptable}{lrrrr}{Mean individual zero-ablation effect among the selected top 50, in nats per prediction. Removed is the percentage of alive standard features excluded by $a_j\geq.01$.}{tab:app-importance}{Model & Removed (\%) & Standard & Post-hoc & Bypass}
GPT-2 & 2.26 & 0.0966 & 0.0130 & 0.0141 \\
Pythia-70M & 3.92 & 0.0372 & 0.0248 & 0.0256 \\
Pythia-160M & 2.36 & 0.0690 & 0.0183 & 0.0182 \\
Pythia-1.4B & 1.04 & 0.0536 & 0.0064 & 0.0059 \\
Pythia-6.9B & 0.74 & 0.0401 & 0.0050 & 0.0044 \\
Llama-3.1-8B & 0.74 & 0.0165 & 0.0070 & 0.0064 \\
Qwen-2.5-7B & 2.08 & 0.1101 & 0.0181 & 0.0100 \\
Gemma-2-2B & 1.57 & 0.0159 & 0.0087 & 0.0110 \\
Gemma-2-9B & 0.72 & 0.0180 & 0.0152 & 0.0177 \\
\end{apptable}

\begin{apptable}{lrrrrr}{Standard-SAE sharing (\%) among the top-$N$ features.}{tab:app-topn}{Model & Dictionary & Top 10 & Top 50 & Top 150 & Top 300}
GPT-2 & 46.6 & 10.0 & 5.5 & 45.8 & 60.1 \\
Pythia-70M & 47.0 & 30.0 & 47.5 & 58.3 & 62.3 \\
Pythia-160M & 49.0 & 22.5 & 13.0 & 48.5 & 63.8 \\
Pythia-1.4B & 62.9 & 20.0 & 6.0 & 52.8 & 69.4 \\
Pythia-6.9B & 62.4 & 20.0 & 4.0 & 55.5 & 71.6 \\
Llama-3.1-8B & 66.2 & 67.5 & 39.5 & 63.2 & 73.1 \\
Qwen-2.5-7B & 47.8 & 10.0 & 6.0 & 39.0 & 62.0 \\
Gemma-2-2B & 58.1 & 100.0 & 87.0 & 83.3 & 81.0 \\
Gemma-2-9B & 58.1 & 97.5 & 92.0 & 90.2 & 84.4 \\
\end{apptable}

For component ablations and subspace comparisons, we group features whose PC1 alignment exceeds five times the median among alive features: $a_j>5\operatorname{median}_{\ell\in A}a_\ell$, where $A$ is the alive set. For uniformly random unit directions, $a_j\sim\operatorname{Beta}(1/2,(d-1)/2)$; this cutoff selects approximately 13\% of directions at these dimensions.

\subsection{Other measures of feature importance}
\label{app:importance_metrics}

Mean ablation replaces a feature's activation with its average on the calibration data. Gradient--activation importance is the sum of $|z_j\,\partial\ell/\partial z_j|$ over valid tokens. Integrated gradients integrate along the straight path from all-zero SAE coefficients to their observed values, then sum absolute attributions over tokens. For bypass, the PC1 coefficient stays fixed along the path.

\begin{apptable}{lrrrr}{Top-50 sharing (\%), standard/bypass.}{tab:app-metrics}{Model & Zero & Mean & Gradient$\times$activation & Integrated gradients}
GPT-2 & 5.5/73.0 & 4.0/71.5 & 12.5/83.5 & 7.0/82.0 \\
Pythia-70M & 47.5/81.0 & 46.0/84.5 & 53.5/64.5 & 44.5/71.0 \\
Pythia-160M & 13.0/83.0 & 10.0/84.5 & 17.0/74.0 & 15.0/78.0 \\
Pythia-1.4B & 6.0/92.0 & 2.0/92.0 & 6.0/89.5 & 10.0/90.5 \\
Pythia-6.9B & 4.0/95.5 & 2.0/93.5 & 7.5/93.5 & 9.0/94.5 \\
Llama-3.1-8B & 39.5/94.5 & 39.5/93.0 & 17.0/93.0 & 11.0/96.5 \\
Qwen-2.5-7B & 6.0/99.5 & 6.0/99.5 & 0.0/94.0 & 0.0$^*$/96.0 \\
Gemma-2-2B & 87.0/93.5 & 86.5/90.5 & 45.5/63.0 & 34.0/86.5 \\
Gemma-2-9B & 92.0/94.5 & 95.0/94.5 & 16.0/46.5 & 31.0$^*$/88.0 \\
\end{apptable}

We approximate the integral with midpoint gradients, starting from 16 intervals and doubling to at most 128. A run is accepted once the attributions sum to within 1\% of the loss change, and a further doubling keeps at least 48 of the top 50 features. Of 92 total runs, 13 do not meet both conditions; the two affected table entries are marked by asterisks.

We check the gradients themselves on GPT-2. Finite-difference estimates in float64 match autograd to a relative error of $10^{-6}$. Furthermore, recomputing the gradient--activation rankings in float32 leaves the top 50 unchanged. An inconsistency should be noted: while the convergence check was applied to the signed sum $|\sum\mathrm{IG}|$, the reported score is $\sum|\mathrm{IG}|$. Rerunning the check on GPT-2 with the reported score keeps 50 top features for the standard SAE and 49 for bypass.

\subsection{Resample ablation}
\label{app:rx_resampling}

\begin{apptable}{lp{.71\linewidth}}{Resample ablation. Each feature uses 16 draws, with replacement values taken from calibration articles separate from the ranking articles.}{tab:app-resample-method}{Method & Replacement rule}
Marginal & Replace every activation with a value sampled from that feature's calibration activations, including zeros. \\
Support-preserving & Leave zeros unchanged; replace positive activations with sampled positive values for the same feature. \\
Position-stratified & Leave zeros unchanged; sample positive values from the same range of token positions. \\
\end{apptable}

Token positions are grouped as 0, 1, 2--3, 4--7, 8--15, 16--31, 32--63, and 64 onward. If a group has fewer than 32 positive calibration values, we sample from all positive calibration values for that feature.

We evaluate all alive features in GPT-2, Pythia-70M, and Pythia-160M. For each other model and SAE type, we evaluate features appearing in any of the top-256 lists for zero ablation, gradient--activation, or integrated gradients, together with the feature cluster defined above and 256 random alive features.

\begin{apptable}{lrrrrrr}{Top-50 sharing (\%), standard/bypass. Counts give evaluated/alive features. Baseline is sharing within the evaluated set; only the first three models evaluate all alive features.}{tab:rx_resampling}{Model & $N_S$ & $N_B$ & Marginal & Support & Position & Baseline S/B}
GPT-2 & 2962/2962 & 3070/3070 & 4.0/77.5 & 83.0/85.0 & 87.0/89.0 & 46.6/44.3 \\
Pythia-70M & 2016/2016 & 2035/2035 & 32.0/80.0 & 51.5/83.0 & 71.0/78.5 & 47.0/45.5 \\
Pythia-160M & 3046/3046 & 3066/3066 & 6.0/76.5 & 50.5/90.0 & 75.0/88.5 & 49.0/46.5 \\
Pythia-1.4B & 1207/6552 & 880/7223 & 0.0/93.5 & 27.5/95.5 & 42.5/97.5 & 64.9/73.3 \\
Pythia-6.9B & 697/9107 & 1138/10209 & 2.0/95.5 & 51.5/94.0 & 74.0/94.0 & 69.4/73.0 \\
Llama-3.1-8B & 1418/9795 & 1001/11996 & 8.0/96.5 & 96.5/97.5 & 97.5/99.5 & 67.0/75.2 \\
Qwen-2.5-7B & 644/5098 & 951/7638 & 0.0/99.5 & 52.0/98.0 & 52.0/99.0 & 55.3/74.4 \\
Gemma-2-2B & 1077/7501 & 1749/8047 & 73.0/93.5 & 90.0/95.5 & 92.5/98.0 & 60.4/64.2 \\
Gemma-2-9B & 872/9603 & 1082/10264 & 92.5/94.5 & 92.5/99.0 & 92.5/100.0 & 65.7/64.0 \\
\end{apptable}

\section{Rank-One Reconstruction and Component Ablations}
\label{app:theory}

\subsection{Non-unique sparse reconstruction}

Take $\mathbf x=a\mathbf v$, where $a\in\mathbb R$, $\|\mathbf v\|_2=1$, and $0<\mathbb E[a^2]<\infty$. Decoder columns have unit norm. Codes are nonnegative, with at most $k$ nonzeros. Biases are zero; encoder weights are unrestricted.

\begin{proposition}[Rank-one reconstruction]
\label{prop:app-rankone}
Assume $d\geq3$, $m\geq4$, and $2\leq k\leq m$. Infinitely many TopK SAEs have zero reconstruction error on $a\mathbf v$. For any $\tau\in(0,1)$, two of them have active decoder sets $\mathcal D,\mathcal D'$ satisfying
\[
\mathbf v\in\operatorname{span}(\mathcal D)\cap\operatorname{span}(\mathcal D'),
\qquad
\max_{\mathbf d\in\mathcal D,\,\mathbf d'\in\mathcal D'}|\mathbf d^\top\mathbf d'|<\tau.
\]
\end{proposition}
\begin{proof}
Extend $\mathbf v$ to an orthonormal triple $(\mathbf v,\mathbf u,\mathbf u')$. Let $\theta\in(0,\pi/2)$ and $a_\pm=\max(\pm a,0)$. The decoder and encoder are
\[
\begin{aligned}
\mathbf d_\pm&=\cos\theta\,\mathbf v\pm\sin\theta\,\mathbf u,\\
D&=(\mathbf d_+,\mathbf d_-,-\mathbf d_+,-\mathbf d_-),\\
W&=\frac{(1,1,-1,-1)^\top\mathbf v^\top}{2\cos\theta}.
\end{aligned}
\]
Pad to width $m$ using zero encoder rows and unit decoder columns. The code and reconstruction are
\[
\begin{aligned}
\mathbf z&=\operatorname{TopK}(\operatorname{ReLU}(W\mathbf x))
=\frac{(a_+,a_+,a_-,a_-,0,\ldots,0)^\top}{2\cos\theta},\\
D\mathbf z&=(a_+-a_-)\mathbf v=\mathbf x.
\end{aligned}
\]
Use $\mathbf u'$ in place of $\mathbf u$ for $D'$. Its columns satisfy
\[
|\mathbf d^\top\mathbf d'|=\cos^2\theta,
\qquad
\mathbf v=\frac{\mathbf d_++\mathbf d_-}{2\cos\theta}
=\frac{\mathbf d'_++\mathbf d'_-}{2\cos\theta}.
\]
Here $\mathbf d,\mathbf d'$ range over the four constructed columns of $D,D'$. Choose $\cos^2\theta<\tau$. Changing $\theta$ gives distinct exact solutions.
\end{proof}

Two columns suffice for a single sign of $a$. The aligned decoder $(\mathbf v,-\mathbf v)$ works for both signs, with code $(a_+,a_-)$.

\begin{proposition}[Magnitude cost]
\label{prop:app-l1}
Let $\mathbf x=\sum_jz_j\mathbf d_j$, with $z_j\geq0$ and $\|\mathbf d_j\|_2=1$. Then
\[
\|\mathbf x\|_2\leq\sum_jz_j.
\]
Equality for $\mathbf x\ne0$ holds iff $\mathbf d_j=\mathbf x/\|\mathbf x\|_2$ whenever $z_j>0$. Above, the total coefficient is $|a|/\cos\theta$.
\end{proposition}
\begin{proof}
For $\mathbf x\ne0$, put $\mathbf e=\mathbf x/\|\mathbf x\|_2$. Then
\[
\|\mathbf x\|_2=\sum_jz_j(\mathbf e^\top\mathbf d_j)\leq\sum_jz_j,
\]
with equality iff $\mathbf e^\top\mathbf d_j=1$ for all $z_j>0$. The bound at $\mathbf x=0$ follows from $z_j\geq0$. In our example, add the two coefficients $|a|/(2\cos\theta)$.
\end{proof}

Squared reconstruction errors after removing features:
\[
\begin{array}{lcc}
\text{Features removed} & \text{Total error} & \text{Error perpendicular to }\mathbf v\\
\hline
\text{None} & 0 & 0\\
\text{Either one} & a^2/(4\cos^2\theta) & a^2\tan^2\theta/4\\
\text{Both} & a^2 & 0
\end{array}
\]
The error in the middle row diverges as $\theta\to\pi/2$ for any fixed $a\ne0$.

\subsection{Individual and joint component ablations}
\label{app:component_ablation}

For contribution $\mathbf q_j=z_j\mathbf d_j$, we separately remove $\mathbf q_j$, $P\mathbf q_j$, or $Q\mathbf q_j$ from the SAE-spliced activation. We use seed 42's top 50 by zero ablation and the same ranking articles for all three interventions (Table~\ref{tab:app-components}).

\begin{apptable}{lrrrr}{Sums of individual top-50 ablation effects, in nats per prediction. The final column is a ratio of sums, not a causal decomposition of loss.}{tab:app-components}{Model & Whole & Parallel & Orthogonal & Orthogonal/whole (\%)}
GPT-2 & 4.834 & 0.055 & 4.713 & 97.5 \\
Pythia-70M & 1.862 & 0.028 & 1.961 & 105.3 \\
Pythia-160M & 3.376 & -0.027 & 3.389 & 100.4 \\
Pythia-1.4B & 2.676 & 0.564 & 2.347 & 87.7 \\
Pythia-6.9B & 1.992 & 0.686 & 1.545 & 77.6 \\
Llama-3.1-8B & 0.825 & 0.138 & 0.589 & 71.3 \\
Qwen-2.5-7B & 5.504 & 0.276 & 5.072 & 92.2 \\
Gemma-2-2B & 0.793 & -0.021 & 0.837 & 105.6 \\
Gemma-2-9B & 0.898 & -0.055 & 0.959 & 106.7 \\
\end{apptable}

Removing only the orthogonal contributions retains most or all of the whole-feature effect, although PC1 reconstruction remains unchanged. Cross-entropy is nonlinear: parallel and orthogonal effects need not add, and either can be negative.

\begin{apptable}{lrrrr}{Joint ablation of the geometric cluster, source seed 42. Effects are nats per prediction. $^*$Qwen is evaluated in bfloat16.}{tab:app-joint}{Model & Features & Whole & Parallel & Orthogonal}
GPT-2 & 455 & 2.385 & 0.825 & 1.318 \\
Pythia-70M & 318 & 1.071 & 0.128 & 1.240 \\
Pythia-160M & 486 & 1.202 & 0.224 & 1.044 \\
Pythia-1.4B & 713 & 1.041 & 0.572 & 0.437 \\
Pythia-6.9B & 114 & 0.588 & 0.421 & 0.120 \\
Llama-3.1-8B & 935 & 2.626 & 2.148 & 0.558 \\
Qwen-2.5-7B$^*$ & 141 & 7.719 & 6.114 & 0.662 \\
Gemma-2-2B & 585 & 0.977 & 0.185 & 0.780 \\
Gemma-2-9B & 352 & 1.193 & 0.348 & 0.831 \\
\end{apptable}

Qwen's joint top-50 ablation overflowed in float16, so we repeated it in bfloat16. Over 33,244 predictions, removing the whole contributions increases loss by 5.652 nats per prediction; removing only the parallel or orthogonal parts gives 4.433 and 1.982.

\subsection{Subspace agreement and cancellation}

Table~\ref{tab:app-subspace} compares the decoder spans of each model's seed-42 cluster and the other four seeds. We take an orthonormal basis for each span and compute the cosines of their principal angles by SVD. We average the squared cosines over angles, then over the four seed comparisons. Independent, uniformly random subspaces of dimensions $p$ and $q$ in $\mathbb R^d$ have expected score $\max(p,q)/d$.

\begin{apptable}{lrrr}{Geometric-cluster subspace agreement. The null uses the ranks of both subspaces in each comparison.}{tab:app-subspace}{Model & Source rank & Mean $\cos^2\theta$ & Isotropic null}
GPT-2 & 455 & 0.683 & 0.592 \\
Pythia-70M & 318 & 0.761 & 0.686 \\
Pythia-160M & 486 & 0.703 & 0.635 \\
Pythia-1.4B & 713 & 0.591 & 0.350 \\
Pythia-6.9B & 114 & 0.274 & 0.029 \\
Llama-3.1-8B & 935 & 0.601 & 0.311 \\
Qwen-2.5-7B & 141 & 0.358 & 0.041 \\
Gemma-2-2B & 585 & 0.548 & 0.272 \\
Gemma-2-9B & 352 & 0.457 & 0.117 \\
\end{apptable}

Cancellation is measured directly in activation space by
\[
 C_J=\frac{\sum_t\sum_{j\in J}\|Qz_{tj}\mathbf d_j\|_2^2}
 {\sum_t\|\sum_{j\in J}Qz_{tj}\mathbf d_j\|_2^2}.
\]
For standard SAEs, $C_J$ ranges from 1.93 in Gemma-2-9B to 456.6 in Qwen; random feature sets of the same size give 0.89--0.99.

We tried 40 random restarts across eight models, training only the cluster for 5,000 steps. None came within 1\% of the held-out reconstruction error of the original cluster given the same number of training steps.

\section{SAE Objectives and Initialization}

\subsection{Magnitude penalties}
\label{app:beta}

The GPT-2 sweep follows Section~\ref{sec:beta}, with minibatch order fixed across the five seeds. We use separate WikiText validation articles for calibration and selection, and the test split for evaluation. We include only features that fire on calibration data. We average sharing over all 20 seed comparisons.

\begin{apptable}{lrrrrrrr}{GPT-2 magnitude penalties, five seeds. Sharing is in percent; content FVU is mean $\pm$ seed SD.}{tab:app-beta}{SAE & $\beta$ & $S_{50}$ sel. & $S_{50}$ test & Dict. & $L_0$ & Total FVU & Content FVU}
Standard & 0.00 & 6.4 & 6.4 & 45.8 & 64.00 & 0.004838 & 0.13941 $\pm$ 0.00034 \\
Standard & 0.03 & 17.9 & 14.8 & 47.9 & 63.97 & 0.004824 & 0.13894 $\pm$ 0.00028 \\
Standard & 0.10 & 49.9 & 52.1 & 51.2 & 63.80 & 0.004820 & 0.13865 $\pm$ 0.00014 \\
Standard & 0.30 & 95.2 & 94.9 & 57.3 & 63.47 & 0.004830 & 0.13869 $\pm$ 0.00022 \\
Bypass & 0.00 & 69.1 & 65.8 & 44.3 & 63.84 & 0.004739 & 0.13666 $\pm$ 0.00021 \\
Bypass & 0.03 & 82.9 & 82.2 & 47.1 & 63.77 & 0.004734 & 0.13644 $\pm$ 0.00011 \\
Bypass & 0.10 & 91.4 & 91.9 & 51.1 & 63.62 & 0.004730 & 0.13633 $\pm$ 0.00013 \\
Bypass & 0.30 & 94.3 & 94.8 & 57.8 & 63.40 & 0.004740 & 0.13662 $\pm$ 0.00014 \\
\end{apptable}

\begin{apptable}{lrrrrrr}{PC1 alignment in the standard GPT-2 penalty sweep. Intermediate: $.01\leq a_j<.3$; high: $a_j\geq.3$; low: $a_j<.01$. Sharing and top-50 fractions are percentages.}{tab:app-beta-geometry}{$\beta$ & Int. count & Int. shared & Int. in top 50 & High count & High shared & Low shared}
0.00 & 65.2 & 2.7 & 93.6 & 1.8 & 95.0 & 46.7 \\
0.03 & 96.6 & 25.9 & 93.2 & 2.4 & 90.0 & 48.6 \\
0.10 & 461.4 & 63.3 & 80.0 & 5.0 & 54.4 & 49.0 \\
0.30 & 1306.6 & 71.7 & 72.8 & 9.8 & 70.9 & 46.3 \\
\end{apptable}

The Pythia-1.4B sweep compares seeds 42 and 137 over a wider range of penalties (Table~\ref{tab:app-pythia-beta}). Reconstruction and exact nonzero counts are reevaluated on the first 100,000 training-store rows.

\begin{apptable}{lrrrrrr}{Pythia-1.4B magnitude penalties. Sharing is in percent.}{tab:app-pythia-beta}{$\beta$ & $S_{50}$ & Dict. & $L_0$: 42 & $L_0$: 137 & Total FVU & Content FVU}
0 & 10.0 & 62.5 & 63.96 & 63.99 & 0.009167 & 0.22767 \\
1e-06 & 10.0 & 62.5 & 63.97 & 64.00 & 0.009163 & 0.22756 \\
1e-05 & 8.0 & 62.6 & 63.98 & 63.99 & 0.009166 & 0.22764 \\
0.0001 & 8.0 & 63.2 & 63.98 & 63.98 & 0.009163 & 0.22756 \\
0.001 & 10.0 & 62.1 & 63.98 & 63.98 & 0.009162 & 0.22753 \\
0.003 & 12.0 & 63.7 & 63.98 & 63.98 & 0.009160 & 0.22748 \\
0.01 & 16.0 & 63.5 & 63.95 & 63.98 & 0.009150 & 0.22721 \\
0.03 & 46.0 & 63.8 & 63.81 & 63.67 & 0.009145 & 0.22700 \\
0.1 & 94.0 & 67.5 & 62.91 & 62.86 & 0.009137 & 0.22656 \\
0.3 & 94.0 & 71.8 & 62.64 & 62.64 & 0.009191 & 0.22764 \\
1 & 98.0 & 79.4 & 62.29 & 62.29 & 0.009448 & 0.23382 \\
3 & 100.0 & 87.6 & 50.21 & 49.25 & 0.011084 & 0.27288 \\
10 & 98.0 & 92.8 & 8.43 & 8.41 & 0.022775 & 0.53503 \\
\end{apptable}

To measure agreement outside PC1, we match each source feature to its best full-decoder match once, then project PC1 out of both directions and renormalize. Pairs in which either remainder is below 0.01 of the original norm are discarded. Results are reported in Table~\ref{tab:app-projected}.

\begin{apptable}{lrrrrrr}{Top-50 matches before and after projecting out PC1. Cosines are means; sharing uses cosine $\geq0.7$ and is in percent.}{tab:app-projected}{Model & $\beta$ & Full cosine & Perp. cosine & Full shared & Perp. shared & Excluded}
GPT-2 & 0 & 0.178 & 0.165 & 6.0 & 4.0 & 0 \\
GPT-2 & 0.03 & 0.316 & 0.235 & 18.0 & 16.0 & 0 \\
GPT-2 & 0.1 & 0.662 & 0.456 & 50.0 & 50.0 & 0 \\
GPT-2 & 0.3 & 0.901 & 0.831 & 96.0 & 88.0 & 0 \\
Pythia-1.4B & 0 & 0.268 & 0.114 & 10.0 & 10.0 & 0 \\
Pythia-1.4B & 0.001 & 0.281 & 0.115 & 10.0 & 10.0 & 0 \\
Pythia-1.4B & 0.01 & 0.295 & 0.219 & 16.0 & 16.0 & 0 \\
Pythia-1.4B & 0.03 & 0.613 & 0.446 & 46.0 & 42.0 & 0 \\
Pythia-1.4B & 0.1 & 0.928 & 0.910 & 94.0 & 92.0 & 0 \\
Pythia-1.4B & 0.3 & 0.916 & 0.911 & 94.0 & 94.0 & 0 \\
Pythia-1.4B & 1 & 0.965 & 0.964 & 98.0 & 98.0 & 0 \\
\end{apptable}

\subsection{Comparison across objectives}
\label{app:archgrid}

The objective comparison uses the same activation stores, width, and training budget as the default TopK runs. It ranks the 1,500 alive features with largest mean squared activation on 64 validation sequences and compares seed 42 with the other seeds.

BatchTopK \citep{bussmann2024batchtopk} keeps the largest $64B$ nonnegative entries in each batch of $B$ tokens and, at inference, uses a threshold tracked during training (decay 0.999). ReLU+$\ell_1$ penalizes the coefficients. Gated \citep{rajamanoharan2024gated} penalizes positive gate preactivations and adds a gate reconstruction loss with a detached decoder. JumpReLU \citep{rajamanoharan2024jump} initializes thresholds from the data and recalibrates them periodically; we call this variant scheduled-threshold JumpReLU. ReLU and Gated penalties are tuned toward $L_0=64$. Tables~\ref{tab:app-grid} and~\ref{tab:app-grid-controls} give feature sharing, sparsity, and penalty settings. S and B denote standard and bypass SAEs.

\begin{apptable}{llrrrrrrr}{Feature sharing (\%) and mean number of active features ($L_0$) across SAE objectives.}{tab:app-grid}{Model & Objective & Seeds & $S_{50}$ S & $S_{50}$ B & Dict. S & Dict. B & $L_0$ S & $L_0$ B}
GPT-2 & TopK & 5/5 & 5.5 & 79.5 & 46.6 & 44.3 & 64.0 & 63.8 \\
GPT-2 & BatchTopK & 5/5 & 17.5 & 64.0 & 42.6 & 45.1 & 63.0 & 63.8 \\
GPT-2 & JumpReLU (sched.) & 5/5 & 11.5 & 66.5 & 42.8 & 46.4 & 65.1 & 66.0 \\
GPT-2 & ReLU+$\ell_1$ & 5/5 & 100.0 & 100.0 & 78.0 & 82.4 & 49.2 & 57.5 \\
GPT-2 & Gated & 3/3 & 11.0 & 92.0 & 41.1 & 51.5 & 56.7 & 59.9 \\
Pythia-160M & TopK & 5/5 & 9.0 & 83.5 & 49.0 & 46.5 & 64.0 & 64.0 \\
Pythia-160M & BatchTopK & 5/5 & 9.5 & 77.0 & 44.5 & 46.0 & 63.2 & 63.8 \\
Pythia-160M & JumpReLU (sched.) & 5/5 & 14.5 & 88.5 & 44.7 & 47.2 & 65.6 & 66.3 \\
Pythia-160M & ReLU+$\ell_1$ & 5/5 & 96.5 & 94.5 & 78.1 & 84.4 & 52.9 & 50.5 \\
Pythia-160M & Gated & 5/5 & 87.0 & 91.0 & 46.4 & 50.6 & 56.3 & 56.4 \\
Pythia-1.4B & TopK & 5/5 & 10.0 & 85.5 & 62.9 & 58.7 & 64.0 & 64.0 \\
Pythia-1.4B & BatchTopK & 5/5 & 14.0 & 87.0 & 62.1 & 54.1 & 63.4 & 64.1 \\
Pythia-1.4B & JumpReLU (sched.) & 5/5 & 12.0 & 89.0 & 62.3 & 53.6 & 65.8 & 66.8 \\
Pythia-1.4B & ReLU+$\ell_1$ & 5/5 & 98.5 & 97.0 & 83.9 & 87.4 & 52.3 & 46.8 \\
Pythia-1.4B & Gated & 5/5 & 86.0 & 94.5 & 58.5 & 61.7 & 62.1 & 55.2 \\
Pythia-6.9B & TopK & 3/3 & 10.0 & 90.0 & 62.2 & 61.0 & 64.0 & 63.9 \\
Pythia-6.9B & BatchTopK & 3/3 & 18.0 & 88.0 & 67.0 & 60.5 & 63.6 & 64.5 \\
Pythia-6.9B & JumpReLU (sched.) & 3/3 & 26.0 & 91.0 & 64.5 & 60.6 & 66.2 & 68.3 \\
Pythia-6.9B & ReLU+$\ell_1$ & 3/3 & 97.0 & 96.0 & 84.5 & 84.3 & 54.6 & 53.4 \\
Pythia-6.9B & Gated & 3/3 & 15.0 & 92.0 & 60.9 & 69.1 & 70.1 & 46.1 \\
Llama-3.1-8B & TopK & 3/3 & 56.0 & 86.0 & 65.9 & 67.8 & 63.7 & 63.4 \\
Llama-3.1-8B & BatchTopK & 3/3 & 62.0 & 91.0 & 70.5 & 69.3 & 64.6 & 65.2 \\
Llama-3.1-8B & JumpReLU (sched.) & 3/3 & 61.0 & 91.0 & 67.8 & 69.6 & 66.2 & 67.1 \\
Llama-3.1-8B & ReLU+$\ell_1$ & 3/3 & 97.0 & 99.0 & 80.0 & 79.8 & 62.8 & 65.0 \\
Llama-3.1-8B & Gated & 3/3 & 99.0 & 99.0 & 68.7 & 64.1 & 57.9 & 60.2 \\
Qwen-2.5-7B & TopK & 3/3 & 6.0 & 98.0 & 46.4 & 61.3 & 64.0 & 64.0 \\
Qwen-2.5-7B & BatchTopK & 3/3 & 33.0 & 95.0 & 56.3 & 63.7 & 63.3 & 63.5 \\
Qwen-2.5-7B & JumpReLU (sched.) & 3/3 & 32.0 & 100.0 & 56.3 & 62.9 & 64.7 & 66.0 \\
Qwen-2.5-7B & ReLU+$\ell_1$ & 3/3 & 97.0 & 99.0 & 69.2 & 80.0 & 73.6 & 51.9 \\
Qwen-2.5-7B & Gated & 3/3 & 47.0 & 97.0 & 27.8 & 64.4 & 53.9 & 54.5 \\
Gemma-2-2B & TopK & 5/5 & 80.5 & 88.0 & 58.0 & 62.1 & 63.9 & 63.8 \\
Gemma-2-2B & BatchTopK & 5/5 & 72.5 & 87.0 & 60.3 & 62.0 & 63.9 & 64.4 \\
Gemma-2-2B & JumpReLU (sched.) & 3/3 & 76.0 & 89.0 & 61.2 & 62.1 & 65.6 & 67.6 \\
Gemma-2-2B & ReLU+$\ell_1$ & 3/3 & 97.0 & 99.0 & 71.3 & 71.7 & 74.0 & 72.8 \\
Gemma-2-2B & Gated & 3/3 & 79.0 & 96.0 & 57.2 & 61.3 & 47.9 & 51.9 \\
Gemma-2-9B & TopK & 3/3 & 90.0 & 87.0 & 58.3 & 60.5 & 64.0 & 63.9 \\
Gemma-2-9B & BatchTopK & 3/3 & 90.0 & 87.0 & 61.5 & 62.6 & 64.3 & 64.6 \\
Gemma-2-9B & JumpReLU (sched.) & 3/3 & 75.0 & 87.0 & 61.0 & 63.1 & 66.4 & 68.5 \\
Gemma-2-9B & ReLU+$\ell_1$ & 3/3 & 95.0 & 94.0 & 72.4 & 71.6 & 69.2 & 68.4 \\
Gemma-2-9B & Gated & 3/3 & 63.0 & 80.0 & 56.3 & 55.2 & 49.8 & 51.2 \\
\end{apptable}

\begin{apptable}{llrrrr}{ReLU and Gated calibration settings. $c$ is the sparsity penalty coefficient. S/B: standard/bypass; $L_0$ ranges are across seeds.}{tab:app-grid-controls}{Model & Objective & $c$ S & $c$ B & $L_0$ range S & $L_0$ range B}
GPT-2 & ReLU+$\ell_1$ & 5.062 & 4.171 & 48.8--49.4 & 57.0--57.8 \\
GPT-2 & Gated & 5.624 & 5.273 & 54.5--61.0 & 59.7--60.0 \\
Pythia-160M & ReLU+$\ell_1$ & 1.155 & 1.138 & 52.4--53.7 & 50.2--51.2 \\
Pythia-160M & Gated & 1.221 & 1.241 & 56.0--56.5 & 56.0--56.8 \\
Pythia-1.4B & ReLU+$\ell_1$ & 3.186 & 2.939 & 52.0--52.7 & 46.7--46.9 \\
Pythia-1.4B & Gated & 3.326 & 3.681 & 60.8--65.8 & 54.5--55.9 \\
Pythia-6.9B & ReLU+$\ell_1$ & 6.427 & 5.639 & 54.2--54.9 & 53.2--53.5 \\
Pythia-6.9B & Gated & 7 & 9.105 & 69.2--71.3 & 45.8--46.3 \\
Llama-3.1-8B & ReLU+$\ell_1$ & 0.7296 & 0.7313 & 62.6--63.0 & 64.9--65.1 \\
Llama-3.1-8B & Gated & 1 & 1 & 57.9--58.0 & 60.0--60.4 \\
Qwen-2.5-7B & ReLU+$\ell_1$ & 4.417 & 4.426 & 73.1--74.1 & 49.8--55.0 \\
Qwen-2.5-7B & Gated & 5.644 & 5.709 & 47.7--57.6 & 54.4--54.7 \\
Gemma-2-2B & ReLU+$\ell_1$ & 9.802 & 9.871 & 73.7--74.1 & 72.4--73.1 \\
Gemma-2-2B & Gated & 16.48 & 15 & 47.3--48.6 & 51.7--52.1 \\
Gemma-2-9B & ReLU+$\ell_1$ & 11.93 & 12.06 & 69.2--69.4 & 68.4--68.5 \\
Gemma-2-9B & Gated & 22.22 & 20.69 & 48.0--52.9 & 51.0--51.4 \\
\end{apptable}

\paragraph{Gated controls.}
The gate penalty gives no gradient to the bias and scale parameters used only to set feature magnitudes. We add $\beta\|\mathbf z\|_1$ to the loss and vary which parameters this penalty updates (Table~\ref{tab:app-gated}).

\begin{apptable}{llrrrr}{GPT-2 Gated SAEs with penalties on feature magnitudes. Sharing is in percent.}{tab:app-gated}{$\beta$ & Parameters updated & $S_{50}$ & Dict. & $L_0$ & Content FVU}
0 & Original gate penalty & 11.0 & 41.1 & 56.76 & 0.15699 \\
0.03 & All coefficient paths & 57.0 & 43.1 & 56.03 & 0.15188 \\
0.10 & All coefficient paths & 92.0 & 50.7 & 55.08 & 0.14712 \\
0.30 & All coefficient paths & 98.0 & 54.7 & 59.88 & 0.14577 \\
1.00 & All coefficient paths & 93.0 & 63.7 & 65.15 & 0.14573 \\
0.10 & Magnitude bias/scale & 89.0 & 49.7 & 54.92 & 0.14704 \\
0.10 & Shared encoder only & 10.0 & 47.8 & 54.97 & 0.14777 \\
\end{apptable}

We raise the gate penalty coefficient from 7 to 8, and observe the sharing rate rising from 15\% to 77\% on Pythia-6.9B (Table~\ref{tab:app-gated-calibration}). The Qwen grid includes a failed seed-0 run with content FVU approximately 2.4.

\begin{apptable}{lrrrrrr}{Pythia-6.9B Gated calibration. Sharing is in percent; content FVU is averaged across checkpoints.}{tab:app-gated-calibration}{Gate coefficient & $S_{50}$ & Dict. & $L_0$: 42 & $L_0$: 137 & $L_0$: 0 & Content FVU}
7.0 & 15.0 & 60.9 & 71.39 & 69.97 & 69.28 & 0.31909 \\
8.0 & 77.0 & 56.1 & 54.96 & 58.66 & 59.48 & 0.32546 \\
\end{apptable}

\paragraph{Learned JumpReLU thresholds.}
We also train JumpReLU with learned thresholds under the same training budget (Table~\ref{tab:app-learned-jump}).

\begin{apptable}{lllrrr}{JumpReLU threshold controls, mean $\pm$ SD across three seeds. CE is the SAE-spliced loss.}{tab:app-learned-jump}{Model & Threshold/arm & Seeds & $L_0$ & Content FVU & Splice CE}
GPT-2 & learned & 3 & $56.99\pm1.41$ & $0.16337\pm0.00086$ & 4.12589 \\
GPT-2 & learned bypass & 3 & $68.35\pm0.56$ & $0.11971\pm0.00049$ & 4.08565 \\
GPT-2 & scheduled & 3 & $65.24\pm0.13$ & $0.14005\pm0.00054$ & 4.09868 \\
Pythia-1.4B & learned & 3 & $59.71\pm1.47$ & $0.23621\pm0.00045$ & 3.20498 \\
\end{apptable}

\subsection{Initialization controls}
\label{app:native}

We train GPT-2 SAEs from ten initializations, using the same minibatches in the same order. We use the settings in Table~\ref{tab:app-training} and clip gradient norms at 1. We rank features on 64 sequences and evaluate them on 256 test sequences. In the pinned condition, the decoder direction aligned with PC1 stays fixed throughout training.

\begin{apptable}{lrrrrrr}{GPT-2 initialization and training controls. Entries are seed medians.}{tab:app-native-geometry}{Training change & $L_0$ & Alive & Tiles & High alignment & Total FVU & Content FVU}
Standard & 64.00 & 2864.0 & 65.0 & 1.0 & 0.00676 & 0.14898 \\
Canonical init. & 63.99 & 2106.0 & 1.0 & 2.0 & 0.00748 & 0.16501 \\
Canonical pinned & 63.98 & 1879.0 & 0.0 & 2.0 & 0.00788 & 0.17362 \\
Half width & 64.00 & 1529.5 & 65.5 & 1.0 & 0.00799 & 0.17624 \\
Double width & 64.00 & 4851.5 & 64.0 & 1.0 & 0.00593 & 0.13083 \\
$k=32$ & 32.00 & 2977.0 & 35.0 & 1.0 & 0.00832 & 0.18334 \\
$k=128$ & 128.00 & 2639.0 & 70.5 & 0.0 & 0.00514 & 0.11343 \\
Bypass & 63.86 & 3072.0 & 0.0 & 0.0 & 0.00655 & 0.14443 \\
$\beta=0.3$ & 0.05 & 53.0 & 0.0 & 4.5 & 0.04643 & 1.00362 \\
\end{apptable}

Table~\ref{tab:app-native-rank} uses CE and KL rankings relative to the unmodified model, all evaluation positions, and all 90 ordered pairs of seeds. We calculate the intervals as follows:

\begin{center}
\small
\begin{tabular}{@{}lp{.65\linewidth}@{}}
\toprule
Setting & Value \\
\midrule
Confidence level & 95\% \\
Bootstrap draws & 2,000 \\
Resampled in each draw & Seeds and 30 selection articles \\
Articles within each draw & Same sample for every seed \\
Kept fixed & Decoders, alive features, and matching threshold \\
\bottomrule
\end{tabular}
\end{center}

\begin{apptable}{lrrrrrr}{GPT-2 sharing (\%) under CE and KL rankings, with joint 95\% bootstrap intervals.}{tab:app-native-rank}{Training & $N$ & Dict. & CE sharing & CE interval & KL sharing & KL interval}
Standard & 10 & 40.0 & 18.8 & 13.2--21.7 & 12.0 & 9.1--17.0 \\
Standard & 50 & 40.0 & 5.5 & 5.0--8.1 & 4.9 & 4.2--5.6 \\
Standard & 100 & 40.0 & 26.4 & 24.7--28.4 & 30.8 & 29.8--32.4 \\
Bypass & 10 & 37.3 & 81.0 & 72.1--85.6 & 92.0 & 87.0--95.5 \\
Bypass & 50 & 37.3 & 65.4 & 58.8--69.1 & 77.0 & 74.1--81.8 \\
Bypass & 100 & 37.3 & 60.4 & 52.6--61.8 & 78.0 & 75.2--79.8 \\
Canonical init. & 10 & 36.2 & 79.7 & 73.2--85.8 & 83.9 & 76.8--89.8 \\
Canonical init. & 50 & 36.2 & 68.0 & 62.9--72.8 & 74.4 & 71.3--79.4 \\
Canonical init. & 100 & 36.2 & 65.0 & 56.5--64.2 & 72.7 & 68.9--74.6 \\
Train after first & 50 & 38.1 & 40.4 & 36.3--43.9 & 35.0 & 31.6--38.8 \\
$\beta=0.3$ & 50 & 1.4 & 0.2 & 0.0--1.4 & 9.0 & 8.4--9.6 \\
\end{apptable}

Excluding the first position from the ranking loss raises standard-SAE top-50 sharing without retraining (Table~\ref{tab:app-native-position}).

\begin{apptable}{llrrrr}{Top-50 sharing (\%) under position exclusion and one-to-one matching; each pair of rows uses the same trained dictionaries.}{tab:app-native-position}{Training & Ranking tokens & CE & CE 1:1 & KL & KL 1:1}
Standard & All & 5.5 & 5.5 & 4.9 & 4.9 \\
Standard & After first & 70.0 & 69.6 & 80.0 & 79.3 \\
Bypass & All & 65.4 & 65.3 & 77.0 & 76.8 \\
Bypass & After first & 65.7 & 65.5 & 77.1 & 76.8 \\
Canonical init. & All & 68.0 & 68.0 & 74.4 & 74.3 \\
Canonical init. & After first & 69.3 & 69.2 & 77.9 & 77.8 \\
\end{apptable}

\subsection{Objective comparison with matched seeds}

For each model in Table~\ref{tab:app-objective-repeat}, we train standard and bypass SAEs with TopK, BatchTopK, and ReLU+$\ell_1$. There are five seeds for each setting. We use the same rule to select 1,500 features, rank them on 64 validation sequences, and measure sharing for every pair of seeds in both directions (20 comparisons).

\begin{apptable}{llrrr}{Sharing (\%) and mean number of active features ($L_0$) for the repeated objective comparison, standard/bypass.}{tab:app-objective-repeat}{Model & Objective & Top 50 & Dictionary & $L_0$}
GPT-2 & TopK & 5.8/79.1 & 46.5/44.0 & 64.00/63.82 \\
GPT-2 & BatchTopK & 15.9/70.0 & 43.2/45.0 & 63.04/63.77 \\
GPT-2 & ReLU+$\ell_1$ & 99.1/99.2 & 75.3/82.1 & 60.02/59.52 \\
Pythia-160M & TopK & 9.3/85.2 & 49.3/46.0 & 63.99/64.00 \\
Pythia-160M & BatchTopK & 14.6/82.9 & 45.2/46.3 & 63.23/63.76 \\
Pythia-160M & ReLU+$\ell_1$ & 96.8/96.8 & 77.0/82.7 & 58.76/59.00 \\
Pythia-1.4B & TopK & 9.2/85.3 & 62.8/58.8 & 63.99/64.00 \\
Pythia-1.4B & BatchTopK & 15.7/90.0 & 62.3/53.6 & 63.35/64.18 \\
Pythia-1.4B & ReLU+$\ell_1$ & 97.8/98.5 & 83.5/86.7 & 58.40/59.67 \\
Pythia-6.9B & TopK & 8.8/89.6 & 62.4/61.1 & 63.99/63.89 \\
Pythia-6.9B & BatchTopK & 17.9/90.5 & 66.2/60.9 & 63.71/64.59 \\
Pythia-6.9B & ReLU+$\ell_1$ & 95.9/96.2 & 83.5/84.2 & 60.06/60.32 \\
Llama-3.1-8B & TopK & 60.4/91.3 & 66.8/67.9 & 63.71/63.37 \\
Llama-3.1-8B & BatchTopK & 58.6/92.2 & 70.3/69.6 & 64.49/65.25 \\
Llama-3.1-8B & ReLU+$\ell_1$ & 98.9/97.6 & 80.5/79.9 & 62.70/61.14 \\
Qwen-2.5-7B & TopK & 6.4/95.0 & 49.9/61.1 & 63.99/63.97 \\
Qwen-2.5-7B & BatchTopK & 32.3/92.9 & 54.7/62.5 & 63.41/63.77 \\
Qwen-2.5-7B & ReLU+$\ell_1$ & 96.7/97.9 & 71.0/72.7 & 59.15/58.36 \\
Gemma-2-2B & TopK & 82.4/88.5 & 58.3/62.3 & 63.91/63.77 \\
Gemma-2-2B & BatchTopK & 78.7/85.8 & 60.4/61.8 & 63.94/64.38 \\
Gemma-2-2B & ReLU+$\ell_1$ & 98.1/98.2 & 73.8/74.3 & 59.58/59.20 \\
Gemma-2-9B & TopK & 83.7/92.2 & 58.1/60.7 & 63.97/63.92 \\
Gemma-2-9B & BatchTopK & 84.0/85.9 & 61.6/62.7 & 64.20/64.59 \\
Gemma-2-9B & ReLU+$\ell_1$ & 94.6/95.2 & 74.3/74.8 & 59.73/59.36 \\
\end{apptable}

\section{Reconstruction Fidelity}

\subsection{Total and content error}
\label{app:fvutable}

For reconstruction error $\mathbf e=\mathbf x-\widehat{\mathbf x}$, let $P=\mathbf v_1\mathbf v_1^\top$ and $Q=I-P$. On the evaluation set,
\begin{align}
 \mathrm{FVU}_{\rm total}
 &=\frac{\sum\|\mathbf e\|_2^2}{\sum\|\mathbf x-\boldsymbol\mu\|_2^2},&
 \mathrm{FVU}_{\parallel}
 &=\frac{\sum\|P\mathbf e\|_2^2}{\sum\|P(\mathbf x-\boldsymbol\mu)\|_2^2},\\
 \mathrm{FVU}_{\perp}
 &=\frac{\sum\|Q\mathbf e\|_2^2}{\sum\|Q(\mathbf x-\boldsymbol\mu)\|_2^2},&
 w_1&=\frac{\sum\|P(\mathbf x-\boldsymbol\mu)\|_2^2}
 {\sum\|\mathbf x-\boldsymbol\mu\|_2^2}.
\end{align}
Since $P\mathbf e$ and $Q\mathbf e$ are orthogonal,
\begin{equation}
 \mathrm{FVU}_{\rm total}
 =w_1\mathrm{FVU}_{\parallel}+(1-w_1)\mathrm{FVU}_{\perp}.
 \label{eq:app-fvu}
\end{equation}
For $\mathrm{FVU}_{\parallel}\approx0$, this gives $\mathrm{FVU}_{\perp}/\mathrm{FVU}_{\rm total}\approx1/(1-w_1)$.

Table~\ref{tab:app-fidelity} gives total and content FVU on 200,000 WikiText-103 test tokens. Content FVU is $\mathrm{FVU}_{\perp}$. S and B mean standard and bypass. For the standard SAE, $R=\mathrm{FVU}_{\perp}/\mathrm{FVU}_{\rm total}$. Gain is the percentage decrease in content FVU.

\begin{apptable}{lrrrrrrrr}{Total and content FVU on WikiText-103 test tokens.}{tab:app-fidelity}{Model & Seeds & $w_1$ (\%) & Total S & Content S & $R$ & Total B & Content B & Gain}
GPT-2 & 5 & 96.850 & 0.004428 & 0.14056 & 31.7 & 0.004330 & 0.13746 & 2.2 \\
Pythia-70M & 5 & 81.918 & 0.020368 & 0.11251 & 5.5 & 0.020168 & 0.11154 & 0.9 \\
Pythia-160M & 5 & 93.096 & 0.010274 & 0.14874 & 14.5 & 0.010123 & 0.14662 & 1.4 \\
Pythia-1.4B & 5 & 96.206 & 0.008595 & 0.22650 & 26.4 & 0.008459 & 0.22294 & 1.6 \\
Pythia-6.9B & 3 & 95.501 & 0.012763 & 0.28363 & 22.2 & 0.012520 & 0.27831 & 1.9 \\
Llama-3.1-8B & 3 & 96.738 & 0.008330 & 0.25510 & 30.6 & 0.007914 & 0.24259 & 4.9 \\
Qwen-2.5-7B & 3 & 99.888 & 0.000370 & 0.32978 & 891.5 & 0.000244 & 0.21791 & 33.9 \\
Gemma-2-2B & 5 & 77.575 & 0.046726 & 0.20827 & 4.5 & 0.046164 & 0.20586 & 1.2 \\
Gemma-2-9B & 3 & 69.286 & 0.066676 & 0.21699 & 3.3 & 0.066122 & 0.21528 & 0.8 \\
\end{apptable}

\subsection{PCA baselines}
\label{app:baselines}

We compare SAEs with fixed and adaptive PCA, each using 64 coefficients outside PC1. Fixed PCA uses the next 64 principal components. Adaptive PCA selects the 64 coefficients with the largest squared values for each token. We fit PCA on one million randomly sampled training activations and evaluate all methods on the final 100,000 training activations, using SAE seed 42 (Table~\ref{tab:app-pca}).

\begin{apptable}{lrrrr}{Content FVU for PCA reconstructions and TopK SAEs.}{tab:app-pca}{Model & Fixed PCA & Adaptive PCA & Standard SAE & Bypass SAE}
GPT-2 & 0.54452 & 0.36167 & 0.14137 & 0.13786 \\
Pythia-70M & 0.33588 & 0.21147 & 0.11594 & 0.11471 \\
Pythia-160M & 0.43916 & 0.30388 & 0.15142 & 0.14952 \\
Pythia-1.4B & 0.66039 & 0.52780 & 0.22899 & 0.22582 \\
Pythia-6.9B & 0.74459 & 0.63860 & 0.28156 & 0.27637 \\
Llama-3.1-8B & 0.69284 & 0.58520 & 0.24665 & 0.23346 \\
Qwen-2.5-7B & 0.59454 & 0.47230 & 0.31584 & 0.21153 \\
Gemma-2-2B & 0.52252 & 0.42584 & 0.20689 & 0.20397 \\
Gemma-2-9B & 0.52175 & 0.43506 & 0.21028 & 0.20839 \\
\end{apptable}

\subsection{Matched coefficient budgets}
\label{app:capacity}

The extra PC1 coefficient gives bypass 65 coefficients in total. We repeat training with $k=63$ to bring this down to 64. Table~\ref{tab:app-budget} gives content FVU and the fraction of the improvement retained, $G_{63}$:
\[
 G_{63}=\frac{\mathrm{FVU}_{\perp,\rm std}-\mathrm{FVU}_{\perp,63+1}}
 {\mathrm{FVU}_{\perp,\rm std}-\mathrm{FVU}_{\perp,64+1}}.
\]
The first three models use the test set above. For Qwen, we use three seeds and 64 validation sequences. The Total column counts PC1; $L_0$ does not.

\begin{apptable}{llrrrrr}{Reconstruction with 64 or 65 total coefficients.}{tab:app-budget}{Model & Method & Seeds & Total & $L_0$ & Content FVU & $G_{63}$ (\%)}
GPT-2 & Standard & 5 & 64 & 64.00 & $0.14056\pm0.00028$ & -- \\
GPT-2 & Bypass 63+1 & 5 & 64 & 62.82 & $0.13822\pm0.00011$ & 75.3 \\
GPT-2 & Bypass 64+1 & 5 & 65 & 63.82 & $0.13746\pm0.00012$ & -- \\
Pythia-1.4B & Standard & 3 & 64 & 63.98 & $0.22642\pm0.00015$ & -- \\
Pythia-1.4B & Bypass 63+1 & 3 & 64 & 63.00 & $0.22364\pm0.00010$ & 80.1 \\
Pythia-1.4B & Bypass 64+1 & 3 & 65 & 64.00 & $0.22295\pm0.00012$ & -- \\
Gemma-2-2B & Standard & 3 & 64 & 63.92 & $0.20849\pm0.00030$ & -- \\
Gemma-2-2B & Bypass 63+1 & 3 & 64 & 62.81 & $0.20678\pm0.00013$ & 67.5 \\
Gemma-2-2B & Bypass 64+1 & 3 & 65 & 63.77 & $0.20597\pm0.00005$ & -- \\
Qwen-2.5-7B & Standard & 3 & 64 & 63.98 & $0.32329\pm0.00839$ & -- \\
Qwen-2.5-7B & Bypass 63+1 & 3 & 64 & 62.95 & $0.20184\pm0.00013$ & 99.4 \\
Qwen-2.5-7B & Bypass 64+1 & 3 & 65 & 63.95 & $0.20109\pm0.00056$ & -- \\
\end{apptable}

Table~\ref{tab:app-allocation} gives counts of PC1-aligned features ($a_j\geq0.01$) and mean squared reconstruction error outside PC1 for large-norm and ordinary tokens. The norm cutoffs are GPT-2: 752.45; Pythia-1.4B: 525.76; Qwen: 2847.50. We use 200,000 tokens per model and average over five, five, and three seeds, respectively. Gain share gives the percentage of the total error reduction due to each group, averaged over seeds.

\begin{apptable}{llrrrrr}{Active PC1-aligned features and reconstruction error by token norm.}{tab:app-allocation}{Model & Tokens & Tokens (\%) & Active $a_j\geq.01$ & Error S & Error B & Gain share (\%)}
GPT-2 & Large norm & 1.40 & 63.11 & 247.22 & 105.95 & 15.9 \\
GPT-2 & Ordinary & 98.60 & 2.85 & 575.84 & 565.08 & 84.1 \\
Pythia-1.4B & Large norm & 2.87 & 62.94 & 158.64 & 61.08 & 47.0 \\
Pythia-1.4B & Ordinary & 97.13 & 1.26 & 385.30 & 382.05 & 53.0 \\
Qwen-2.5-7B & Large norm & 1.88 & 63.17 & 21003.88 & 2506.10 & 66.8 \\
Qwen-2.5-7B & Ordinary & 98.12 & 26.91 & 1166.54 & 988.66 & 33.2 \\
\end{apptable}

\subsection{Splice cost}
\label{app:splice}

Splice cost is the increase in cross-entropy when model activations are replaced by their SAE reconstructions, including restored PC1 for bypass. Table~\ref{tab:app-splice} uses seed 42 and 256 validation sequences per model. Model CE is cross-entropy before replacement; reduction is the percentage decrease in splice cost from standard to bypass.

\begin{apptable}{lrrrrr}{Cross-entropy and its increase under SAE reconstruction, in nats per prediction.}{tab:app-splice}{Model & Predictions & Model CE & Standard cost & Bypass cost & Reduction (\%)}
GPT-2 & 17,558 & 3.95775 & 0.08054 & 0.07386 & 8.3 \\
Pythia-70M & 17,571 & 4.53997 & 0.17583 & 0.17470 & 0.6 \\
Pythia-160M & 17,571 & 3.95300 & 0.15933 & 0.14691 & 7.8 \\
Pythia-1.4B & 17,571 & 3.19029 & 0.10813 & 0.09693 & 10.4 \\
Pythia-6.9B & 17,571 & 2.96687 & 0.23805 & 0.21291 & 10.6 \\
Llama-3.1-8B & 17,931 & 2.72509 & 0.08931 & 0.08647 & 3.2 \\
Qwen-2.5-7B & 18,059 & 2.50963 & 0.11548 & 0.06485 & 43.8 \\
Gemma-2-2B & 18,097 & 3.10721 & 0.08857 & 0.08438 & 4.7 \\
Gemma-2-9B & 18,097 & 2.88440 & 0.08041 & 0.07812 & 2.8 \\
\end{apptable}

\section{Behavioral Interventions}
\label{app:behavioral}

\subsection{Subject--verb agreement prompts}

We construct 32,000 prompts from ten sentence templates, 40 subject words, and 20 distractor words, each in singular and plural form. Changing the subject gives the donor prompt; changing the distractor gives the control. The donor requires a different verb, while the control keeps the original answer. We keep examples of equal token length with corresponding words at the same positions. Calibration, feature selection, validation, test, and confirmation use separate templates and words.

For each SAE, we select from four sets of 50 features: the standard top 50, the top 50 after excluding $a_j\geq0.01$ (post-hoc), the bypass top 50, and a random set of active features. We rank features by zero-ablation damage on WikiText, then choose edits on the selection examples. At each step, we add the feature that most increases the logit difference between the donor verb and the original verb. For the random comparison, we repeat with three sets and average.

\subsection{Editing activations}

Let $\mathbf h_r$ be the original prompt's activation and $\mathbf h_d$ the donor's. For selected features $J$, we apply
\begin{equation}
 \mathbf h'_r=\mathbf h_r+
 \lambda\sum_{j\in J}[z_j(\mathbf h_d)-z_j(\mathbf h_r)]\mathbf d_j.
 \label{eq:app-donor}
\end{equation}
For the control, use $\mathbf h_f$ in place of $\mathbf h_d$. These edits leave the reconstruction error and bypass PC1 coefficient in place.

We edit the final token for Pythia-160M and all non-padding tokens for the other models.

Table~\ref{tab:app-behavior} uses five features and $\lambda=1$. Success means that copying from the donor changes the preferred verb, while copying from the control does not. We exclude examples that already favor the donor verb. Control is the percentage of all test prompts whose preferred verb is unchanged by the control edit, averaged over seeds.

\begin{apptable}{llrrrrrr}{Agreement-edit success (\%) with five features and $\lambda=1$.}{tab:app-behavior}{Model & Set & Eligible/total & 42 & 137 & 0 & Mean & Control}
GPT-2 & Standard & 512/512 & 0.00 & 0.00 & 0.00 & 0.00 & 100.00 \\
GPT-2 & Post-hoc & 512/512 & 6.25 & 17.58 & 3.32 & 9.05 & 100.00 \\
GPT-2 & Bypass & 512/512 & 4.10 & 15.23 & 1.95 & 7.10 & 100.00 \\
GPT-2 & Random & 512/512 & 0.00 & 0.00 & 0.00 & 0.00 & 100.00 \\
Pythia-160M & Standard & 264/264 & 0.00 & 0.00 & 0.00 & 0.00 & 100.00 \\
Pythia-160M & Post-hoc & 264/264 & 0.00 & 0.00 & 0.00 & 0.00 & 99.75 \\
Pythia-160M & Bypass & 264/264 & 0.00 & 0.00 & 0.00 & 0.00 & 99.87 \\
Pythia-160M & Random & 264/264 & 0.00 & 0.00 & 0.00 & 0.00 & 100.00 \\
Pythia-1.4B & Standard & 264/264 & 0.00 & 0.00 & 0.00 & 0.00 & 100.00 \\
Pythia-1.4B & Post-hoc & 264/264 & 34.47 & 32.20 & 29.17 & 31.94 & 100.00 \\
Pythia-1.4B & Bypass & 264/264 & 39.39 & 36.74 & 36.36 & 37.50 & 100.00 \\
Pythia-1.4B & Random & 264/264 & 0.00 & 0.00 & 0.00 & 0.00 & 100.00 \\
Qwen-2.5-7B & Standard & 506/512 & 0.00 & 0.00 & 0.00 & 0.00 & 100.00 \\
Qwen-2.5-7B & Post-hoc & 506/512 & 3.56 & 3.75 & 5.34 & 4.22 & 98.31 \\
Qwen-2.5-7B & Bypass & 506/512 & 2.37 & 2.37 & 1.38 & 2.04 & 98.24 \\
Qwen-2.5-7B & Random & 506/512 & 0.00 & 0.00 & 0.00 & 0.00 & 100.00 \\
Gemma-2-2B & Standard & 512/512 & 64.45 & 64.06 & 65.82 & 64.78 & 100.00 \\
Gemma-2-2B & Post-hoc & 512/512 & 64.84 & 63.87 & 65.82 & 64.84 & 100.00 \\
Gemma-2-2B & Bypass & 512/512 & 62.70 & 65.43 & 64.26 & 64.13 & 100.00 \\
Gemma-2-2B & Random & 512/512 & 0.00 & 0.00 & 0.00 & 0.00 & 100.00 \\
\end{apptable}

\subsection{Number of features, edit strength, and transfer}

Table~\ref{tab:app-behavior-sensitivity} gives success rates for $b$ edited features and strength $\lambda$. We vary $b$ with $\lambda=1$ and vary $\lambda$ with $b=5$. Here $\lambda=1$ copies the donor's coefficients; $\lambda=2$ doubles the change.

\begin{apptable}{llrrrrrr}{Agreement-edit success (\%) by number of features and edit strength.}{tab:app-behavior-sensitivity}{Model & Set & $b=1$ & $b=3$ & $b=5$ & $\lambda=.25$ & $\lambda=.5$ & $\lambda=2$}
GPT-2 & Standard & 0.00 & 0.00 & 0.00 & 0.00 & 0.00 & 0.00 \\
GPT-2 & Post-hoc & 8.66 & 8.98 & 9.05 & 0.00 & 0.00 & 49.41 \\
GPT-2 & Bypass & 7.03 & 7.10 & 7.10 & 0.00 & 0.00 & 48.70 \\
GPT-2 & Random & 0.00 & 0.00 & 0.00 & 0.00 & 0.00 & 0.00 \\
Pythia-160M & Standard & 0.00 & 0.00 & 0.00 & 0.00 & 0.00 & 0.00 \\
Pythia-160M & Post-hoc & 0.00 & 0.00 & 0.00 & 0.00 & 0.00 & 0.00 \\
Pythia-160M & Bypass & 0.00 & 0.00 & 0.00 & 0.00 & 0.00 & 0.00 \\
Pythia-160M & Random & 0.00 & 0.00 & 0.00 & 0.00 & 0.00 & 0.00 \\
Pythia-1.4B & Standard & 0.00 & 0.00 & 0.00 & 0.00 & 0.00 & 0.00 \\
Pythia-1.4B & Post-hoc & 31.82 & 31.94 & 31.94 & 0.00 & 0.00 & 72.85 \\
Pythia-1.4B & Bypass & 37.50 & 37.63 & 37.50 & 0.00 & 0.63 & 72.35 \\
Pythia-1.4B & Random & 0.00 & 0.00 & 0.00 & 0.00 & 0.00 & 0.00 \\
Qwen-2.5-7B & Standard & 0.00 & 0.00 & 0.00 & 0.00 & 0.00 & 0.53 \\
Qwen-2.5-7B & Post-hoc & 4.22 & 4.22 & 4.22 & 0.00 & 0.13 & 47.36 \\
Qwen-2.5-7B & Bypass & 1.84 & 1.98 & 2.04 & 0.00 & 0.07 & 29.58 \\
Qwen-2.5-7B & Random & 0.00 & 0.00 & 0.00 & 0.00 & 0.00 & 0.00 \\
Gemma-2-2B & Standard & 64.45 & 64.71 & 64.78 & 0.00 & 0.26 & 78.78 \\
Gemma-2-2B & Post-hoc & 64.45 & 64.84 & 64.84 & 0.00 & 0.33 & 78.65 \\
Gemma-2-2B & Bypass & 64.13 & 64.00 & 64.13 & 0.00 & 0.39 & 78.39 \\
Gemma-2-2B & Random & 0.00 & 0.00 & 0.00 & 0.00 & 0.00 & 0.00 \\
\end{apptable}

Table~\ref{tab:app-transfer} gives success after matching the selected features to another SAE by signed decoder cosine. The edit strength stays the same. Top 50 restricts matching to the other SAE's top 50. Full dict. searches all features; Full + filter then excludes matches outside the top 50. A dash means all matches were excluded. Results average six ordered seed pairs. Original gives success before transfer; Cosine and Repeats give the mean cosine and number of duplicate matches when searching the top 50.

\begin{apptable}{llrrrrrr}{Agreement-edit success (\%) after transferring five-feature edits between SAEs.}{tab:app-transfer}{Model & Set & Original & Top 50 & Full dict. & Full + filter & Cosine & Repeats}
GPT-2 & Standard & 0.00 & 0.00 & 0.00 & 0.00 & 0.126 & 0.17 \\
GPT-2 & Post-hoc & 9.05 & 8.59 & 8.85 & 8.72 & 0.827 & 0.50 \\
GPT-2 & Bypass & 7.10 & 7.23 & 7.75 & 7.23 & 0.888 & 0.17 \\
GPT-2 & Random & 0.00 & 0.00 & 0.00 & 0.00 & 0.127 & 0.22 \\
Pythia-160M & Standard & 0.00 & 0.00 & 0.00 & 0.00 & 0.297 & 0.00 \\
Pythia-160M & Post-hoc & 0.00 & 0.00 & 0.00 & 0.00 & 0.965 & 0.33 \\
Pythia-160M & Bypass & 0.00 & 0.00 & 0.00 & 0.00 & 0.860 & 0.33 \\
Pythia-160M & Random & 0.00 & 0.00 & 0.00 & 0.00 & 0.105 & 0.22 \\
Pythia-1.4B & Standard & 0.00 & 0.00 & 0.00 & 0.00 & 0.269 & 0.67 \\
Pythia-1.4B & Post-hoc & 31.94 & 31.94 & 31.94 & 31.94 & 0.961 & 0.33 \\
Pythia-1.4B & Bypass & 37.50 & 37.25 & 37.18 & 37.50 & 0.943 & 0.00 \\
Pythia-1.4B & Random & 0.00 & 0.00 & 0.00 & -- & 0.085 & 0.11 \\
Qwen-2.5-7B & Standard & 0.00 & 0.00 & 0.00 & 0.00 & 0.446 & 0.33 \\
Qwen-2.5-7B & Post-hoc & 4.22 & 4.22 & 4.22 & 4.22 & 0.967 & 0.00 \\
Qwen-2.5-7B & Bypass & 2.04 & 1.94 & 1.91 & 1.94 & 0.807 & 0.33 \\
Qwen-2.5-7B & Random & 0.00 & 0.00 & 0.00 & 0.00 & 0.088 & 0.28 \\
Gemma-2-2B & Standard & 64.78 & 64.45 & 64.45 & 64.45 & 0.955 & 0.17 \\
Gemma-2-2B & Post-hoc & 64.84 & 64.84 & 64.78 & 64.84 & 0.856 & 0.00 \\
Gemma-2-2B & Bypass & 64.13 & 63.90 & 64.00 & 64.03 & 0.955 & 0.33 \\
Gemma-2-2B & Random & 0.00 & 0.00 & 0.00 & 0.00 & 0.096 & 0.33 \\
\end{apptable}

\subsection{Agreement and indirect-object identification}
\label{app:expanded_patching}

We also test subject--verb agreement (SVA) and indirect-object identification (IOI) with edits applied to SAE reconstructions. The calibration, selection, validation, and test sets have separate templates and words, with at most 512, 2,048, 1,024, and 4,096 examples.

Let $y_r$ and $y_d$ be the original and donor answers, and $\ell_r$ and $\ell_d$ their unedited logit vectors. We keep examples satisfying
\[
 \ell_r(y_r)>\ell_r(y_d),\qquad
 \ell_d(y_d)>\ell_d(y_r).
\]
We require at least 60\% calibration accuracy and 200 eligible examples, which excludes Pythia-70M from IOI.

We use the same three top-50 sets and three random sets as above. On the first 256 eligible selection examples, we add whichever feature gives the highest success rate. Ties are broken by the mean difference between donor-answer and original-answer logits. Edits apply at every non-padding position:
\begin{equation}
 \widehat{\mathbf h}'_r=\widehat{\mathbf h}_r+
 \lambda\sum_{j\in J}[z_j(\mathbf h_d)-z_j(\mathbf h_r)]\mathbf d_j.
 \label{eq:app-spliced-patch}
\end{equation}
Here $\widehat{\mathbf h}_r$ is the SAE reconstruction, including the original prompt's PC1 component for bypass. Success requires the donor edit to favor the donor answer and the control edit to favor the original answer. Table~\ref{tab:app-expanded-patching} reports the mean success rate across seeds with five features and $\lambda=1$. We also test one, three, five, and ten features and $\lambda\in\{0.25,0.5,1,2\}$.

\begin{apptable}{llrrrr}{Agreement and indirect-object edit success (\%).}{tab:app-expanded-patching}{Task & Model & Seeds & Standard & Post-hoc & Bypass}
SVA & GPT-2 & 5 & 0.00 & 34.27 & 26.40 \\
SVA & Pythia-70M & 5 & 74.85 & 74.60 & 76.31 \\
SVA & Pythia-160M & 5 & 0.02 & 15.55 & 17.60 \\
SVA & Pythia-1.4B & 5 & 0.00 & 67.04 & 67.14 \\
SVA & Pythia-6.9B & 1 & 0.00 & 0.00 & 0.00 \\
SVA & Llama-3.1-8B & 1 & 31.85 & 29.82 & 31.97 \\
SVA & Qwen-2.5-7B & 1 & 0.00 & 17.41 & 9.63 \\
SVA & Gemma-2-2B & 5 & 77.14 & 77.09 & 75.10 \\
SVA & Gemma-2-9B & 1 & 56.26 & 56.26 & 63.95 \\
IOI & GPT-2 & 5 & 0.44 & 12.26 & 9.90 \\
IOI & Pythia-160M & 5 & 0.74 & 1.23 & 1.09 \\
IOI & Pythia-1.4B & 5 & 0.21 & 0.42 & 0.60 \\
IOI & Pythia-6.9B & 1 & 1.23 & 3.29 & 2.99 \\
IOI & Llama-3.1-8B & 1 & 0.27 & 1.15 & 1.27 \\
IOI & Qwen-2.5-7B & 1 & 0.02 & 0.93 & 1.50 \\
IOI & Gemma-2-2B & 5 & 2.73 & 3.20 & 2.16 \\
IOI & Gemma-2-9B & 1 & 2.42 & 2.42 & 2.49 \\
\end{apptable}

Some runs retain slightly different test examples. For Table~\ref{tab:app-paired-patching}, we use seed 42 and keep only examples shared by the compared runs. Its 95\% intervals come from 2,000 bootstrap samples of these examples, sampled within each sentence-template group.

\begin{apptable}{llrrr}{Differences in edit success on the same test examples, in percentage points [95\% interval].}{tab:app-paired-patching}{Task & Model & Examples & Post-hoc $-$ standard & Bypass $-$ standard}
SVA & GPT-2 & 4096 & 35.1 [33.6, 36.5] & 20.0 [18.8, 21.2] \\
SVA & Pythia-70M & 3346 & 0.9 [0.1, 1.6] & 5.1 [4.0, 6.2] \\
SVA & Pythia-160M & 3454 & 7.9 [7.0, 8.8] & 20.4 [19.2, 21.8] \\
SVA & Pythia-1.4B & 3456 & 65.2 [63.6, 66.8] & 66.8 [65.2, 68.3] \\
SVA & Pythia-6.9B & 3456 & 0.0 [0.0, 0.0] & 0.0 [0.0, 0.0] \\
SVA & Llama-3.1-8B & 4094 & -2.0 [-2.5, -1.6] & 0.1 [-0.8, 1.0] \\
SVA & Qwen-2.5-7B & 4060 & 17.4 [16.4, 18.4] & 9.6 [8.8, 10.5] \\
SVA & Gemma-2-2B & 4093 & 0.0 [0.0, 0.1] & -0.6 [-1.5, 0.2] \\
SVA & Gemma-2-9B & 4092 & 0.0 [0.0, 0.0] & 7.7 [6.7, 8.8] \\
IOI & GPT-2 & 3246 & 10.3 [9.2, 11.3] & 11.2 [10.1, 12.3] \\
IOI & Pythia-160M & 3561 & 0.1 [-0.3, 0.5] & -0.2 [-0.7, 0.2] \\
IOI & Pythia-1.4B & 4094 & -0.2 [-0.4, 0.0] & -0.2 [-0.5, 0.0] \\
IOI & Pythia-6.9B & 4068 & 2.1 [1.5, 2.6] & 1.7 [1.2, 2.3] \\
IOI & Llama-3.1-8B & 4093 & 0.9 [0.5, 1.2] & 1.0 [0.7, 1.4] \\
IOI & Qwen-2.5-7B & 4077 & 0.9 [0.6, 1.2] & 1.5 [1.1, 1.8] \\
IOI & Gemma-2-2B & 4095 & 0.0 [-0.1, 0.2] & -0.5 [-1.1, 0.1] \\
IOI & Gemma-2-9B & 4096 & 0.0 [0.0, 0.0] & 0.1 [-0.4, 0.6] \\
\end{apptable}

Controls include no edit, copying a prompt's own feature values, replacing the full activation, replacing only PC1, and replacing only the component outside PC1. In IOI examples, the original and donor PC1 coefficients can be identical.

\section{Automated Interpretability}
\label{app:rx_autointerp}

\paragraph{Feature and context selection.}
We evaluate the seed-42 SAEs at the layers in Table~\ref{tab:app-models}. For each SAE, we take the 100 features ranked highest by zero-ablation damage and a random sample of 100 active features. The explanation and evaluation articles are separate, with about 500,000 GPT-2 tokens in each set. Examples are selected as follows:
\begin{center}
\begin{tabular}{p{0.22\linewidth}p{0.70\linewidth}}
\toprule
Use & Examples per feature \\
\midrule
Explanation & Up to ten top-activating examples and five from each quartile of nonzero activations. \\
Evaluation & Up to 50 activating examples across those quartiles and 50 random inactive examples. \\
Harder negatives & Up to 50 inactive examples where a feature with a similar decoder fires. \\
\bottomrule
\end{tabular}
\end{center}
Quartiles are fitted on the explanation articles. An example contains the target token and up to 32 preceding tokens.

\paragraph{Generating and testing explanations.}
Qwen2.5-32B-Instruct and Mistral-Small-24B-Instruct-2501 each write descriptions and test the other's descriptions. The prompt asks for one sentence; generation is greedy with a 120-token limit.

We then test how well the other model predicts feature firing from just the description and a text excerpt. For each prediction, we take the log odds of Yes over No, combining the first-token forms of each answer. Tables~\ref{tab:rx_auto_qwen} and~\ref{tab:rx_auto_mistral} give mean AUROC. Top denotes the top 100 features; Control denotes the random 100. Each pair is standard/bypass.

\begin{apptable}{lrrrr}{Feature-detection AUROC with Qwen explaining and Mistral evaluating.}{tab:rx_auto_qwen}{Model & Top: random & Control: random & Top: hard & Control: hard}
GPT-2 & 0.470/0.727 & 0.753/0.757 & 0.411/0.633 & 0.677/0.679 \\
Pythia-160M & 0.566/0.737 & 0.723/0.737 & 0.539/0.618 & 0.641/0.657 \\
Pythia-1.4B & 0.551/0.732 & 0.773/0.755 & 0.475/0.621 & 0.668/0.646 \\
Pythia-6.9B & 0.568/0.757 & 0.812/0.799 & 0.527/0.636 & 0.700/0.694 \\
Qwen-2.5-7B & 0.521/0.722 & 0.789/0.818 & 0.509/0.629 & 0.670/0.705 \\
Llama-3.1-8B & 0.835/0.742 & 0.819/0.805 & 0.780/0.599 & 0.675/0.665 \\
Gemma-2-2B & 0.800/0.760 & 0.832/0.806 & 0.679/0.623 & 0.708/0.695 \\
Gemma-2-9B & 0.761/0.768 & 0.833/0.801 & 0.620/0.602 & 0.675/0.680 \\
\end{apptable}

\begin{apptable}{lrrrr}{Feature-detection AUROC with Mistral explaining and Qwen evaluating.}{tab:rx_auto_mistral}{Model & Top: random & Control: random & Top: hard & Control: hard}
GPT-2 & 0.665/0.715 & 0.714/0.720 & 0.629/0.619 & 0.672/0.648 \\
Pythia-160M & 0.626/0.708 & 0.704/0.704 & 0.589/0.615 & 0.638/0.661 \\
Pythia-1.4B & 0.669/0.743 & 0.734/0.755 & 0.554/0.631 & 0.665/0.663 \\
Pythia-6.9B & 0.645/0.737 & 0.791/0.766 & 0.594/0.608 & 0.713/0.684 \\
Qwen-2.5-7B & 0.586/0.699 & 0.764/0.779 & 0.625/0.621 & 0.673/0.700 \\
Llama-3.1-8B & 0.831/0.733 & 0.783/0.772 & 0.779/0.581 & 0.691/0.689 \\
Gemma-2-2B & 0.789/0.751 & 0.786/0.757 & 0.684/0.629 & 0.705/0.705 \\
Gemma-2-9B & 0.728/0.757 & 0.777/0.746 & 0.626/0.629 & 0.675/0.666 \\
\end{apptable}

\subsection{Uncertainty}

Table~\ref{tab:app-auto-ci} compares the top features using random negatives. We compute 95\% intervals from 2,000 bootstrap samples of features, sampling separately from features in the top set only, the random set only, and both sets. A feature belonging to both is included in both averages whenever sampled. Standard and bypass features are sampled independently.

\begin{apptable}{lrr}{Bypass minus standard AUROC for the top features, with 95\% intervals.}{tab:app-auto-ci}{Model & Qwen explains & Mistral explains}
GPT-2 & 0.257 [0.196, 0.313] & 0.049 [0.008, 0.092] \\
Pythia-160M & 0.171 [0.119, 0.225] & 0.082 [0.037, 0.123] \\
Pythia-1.4B & 0.181 [0.123, 0.235] & 0.074 [0.030, 0.117] \\
Pythia-6.9B & 0.188 [0.139, 0.237] & 0.093 [0.049, 0.138] \\
Llama-3.1-8B & -0.093 [-0.139, -0.048] & -0.097 [-0.147, -0.047] \\
Qwen-2.5-7B & 0.200 [0.136, 0.267] & 0.113 [0.064, 0.158] \\
Gemma-2-2B & -0.040 [-0.079, -0.001] & -0.038 [-0.079, 0.004] \\
Gemma-2-9B & 0.007 [-0.030, 0.045] & 0.030 [-0.015, 0.074] \\
\end{apptable}

\section{Sparse Probing}
\label{app:rx_probing}

\paragraph{Data and fitting.}
We train classifiers on SAE features for SST-2, AG News, Amazon Polarity, and Emotion, using five standard and five bypass SAEs for each model in Table~\ref{tab:rx_probe20} (seeds 42, 137, 0, 1, and 2). For each task, we randomly sample up to 2,000 training examples and 500 test examples per class. We remove duplicate texts, including test texts that appear in training. Test examples come from each dataset's official held-out split; for SST-2, we use the labelled validation split.

We use logistic regression with $k\in\{1,20,100\}$ nonconstant features. Features are chosen by mutual information between firing and class label. We choose $C\in\{10^{-3},10^{-2},10^{-1},1,10,100\}$ by five-fold cross-validation, repeating feature selection and standardization on each training fold. The selected classifier is refitted on the training set before testing.

Inputs are mean feature activations over at most 64 non-padding tokens, or the final token's activations in inputs of at most 128 tokens.

Table~\ref{tab:rx_probe20} gives accuracy with 20 features, averaged over five seeds. Each pair is standard/bypass. Table~\ref{tab:app-probe-ci} gives their mean difference across tasks ($\Delta$), with 95\% intervals from resampling test examples. Table~\ref{tab:app-probe-budgets} compares the feature counts and choice of tokens.

\begin{apptable}{lrrrrr}{Test accuracy (\%) using 20 SAE features averaged across tokens.}{tab:rx_probe20}{Model & SST-2 & AG News & Amazon & Emotion & $\Delta$}
GPT-2 & 80.78/79.22 & 80.84/80.61 & 79.58/80.24 & 59.35/59.22 & -0.31 \\
Pythia-160M & 78.12/78.35 & 80.92/79.91 & 79.76/80.62 & 55.84/57.76 & +0.50 \\
Pythia-1.4B & 85.69/85.41 & 80.63/83.76 & 86.60/86.96 & 62.10/62.67 & +0.95 \\
Pythia-6.9B & 87.82/87.64 & 80.00/78.44 & 86.98/88.00 & 61.24/63.24 & +0.32 \\
Qwen-2.5-7B & 87.71/88.19 & 83.17/81.49 & 85.06/86.82 & 59.76/60.70 & +0.38 \\
Llama-3.1-8B & 88.83/90.05 & 77.42/76.97 & 88.80/89.80 & 62.70/60.27 & -0.17 \\
Gemma-2-2B & 90.50/90.60 & 80.99/80.56 & 89.56/89.86 & 61.03/62.10 & +0.26 \\
Gemma-2-9B & 91.19/90.02 & 79.15/80.17 & 89.64/89.76 & 59.24/60.33 & +0.26 \\
\end{apptable}

\begin{apptable}{lrr}{Bypass minus standard accuracy with 20 features, in percentage points.}{tab:app-probe-ci}{Model & Difference & 95\% interval}
GPT-2 & -0.31 & [-0.78, 0.17] \\
Pythia-160M & 0.50 & [0.03, 1.00] \\
Pythia-1.4B & 0.95 & [0.54, 1.36] \\
Pythia-6.9B & 0.32 & [-0.10, 0.75] \\
Llama-3.1-8B & -0.17 & [-0.61, 0.28] \\
Qwen-2.5-7B & 0.38 & [-0.12, 0.89] \\
Gemma-2-2B & 0.26 & [-0.15, 0.65] \\
Gemma-2-9B & 0.26 & [-0.18, 0.67] \\
\end{apptable}

\begin{apptable}{lrrrr}{Mean test accuracy (\%) by number of features and choice of tokens.}{tab:app-probe-budgets}{Model & Mean: $k=1$ & Mean: $k=20$ & Mean: $k=100$ & Last: $k=20$}
GPT-2 & 59.10/56.55 & 75.14/74.82 & 78.99/79.13 & 63.49/63.75 \\
Pythia-160M & 55.61/55.52 & 73.66/74.16 & 77.82/78.19 & 55.92/61.78 \\
Pythia-1.4B & 56.43/58.16 & 78.75/79.70 & 82.63/82.98 & 65.11/64.84 \\
Pythia-6.9B & 59.67/59.13 & 79.01/79.33 & 82.96/83.37 & 66.38/66.94 \\
Llama-3.1-8B & 58.20/58.82 & 79.44/79.27 & 83.28/82.89 & 71.18/70.66 \\
Qwen-2.5-7B & 58.19/59.96 & 78.92/79.30 & 81.99/82.89 & 67.72/69.90 \\
Gemma-2-2B & 59.22/58.50 & 80.52/80.78 & 83.90/84.27 & 71.50/71.01 \\
Gemma-2-9B & 62.73/60.79 & 79.81/80.07 & 83.91/83.88 & 69.11/68.82 \\
\end{apptable}

Table~\ref{tab:app-probe-baselines} compares probes using the full residual activation, the activation with PC1 removed, PC1 alone, and 20 randomly chosen active SAE features. Random-feature results average five selections per feature count. Bypass + PC1 adds the PC1 coefficient to 20 selected bypass features. All use mean activations over the first 64 tokens; accuracy is averaged over tasks and SAE seeds where applicable.

\begin{apptable}{lrrrrr}{Mean test accuracy (\%) for probe baselines.}{tab:app-probe-baselines}{Model & Dense & Dense $\perp$ & PC1 & Random S/B & Bypass + PC1}
GPT-2 & 80.02 & 80.04 & 38.19 & 47.78/48.60 & 74.77 \\
Pythia-160M & 79.19 & 79.24 & 38.57 & 48.03/48.25 & 74.16 \\
Pythia-1.4B & 83.89 & 83.87 & 38.71 & 46.79/47.76 & 79.83 \\
Pythia-6.9B & 85.98 & 85.97 & 39.66 & 45.70/45.85 & 79.49 \\
Llama-3.1-8B & 86.41 & 86.40 & 38.38 & 45.54/46.20 & 79.37 \\
Qwen-2.5-7B & 86.20 & 86.27 & 38.72 & 46.65/46.72 & 79.22 \\
Gemma-2-2B & 86.04 & 86.04 & 39.66 & 48.25/47.00 & 80.86 \\
Gemma-2-9B & 86.40 & 86.57 & 39.75 & 47.07/46.24 & 80.27 \\
\end{apptable}

\section{What PC1 Encodes}
\label{app:pc1_characterization}

\subsection{Correlations with norm and token frequency}

For Pythia-1.4B at layer 12, we use 65,535 activations from 87 held-out articles. Table~\ref{tab:app-pc1-associations} reports the absolute Pearson correlation between the centered PC1 coefficient and activation norm, and the Spearman correlation between its absolute value and token frequency. Frequencies are counted in the five-million-token training sample. The 95\% intervals come from 200 bootstrap samples of articles. The absolute cosine between PC1 and the training mean is 0.949.

\begin{apptable}{lrrrrr}{Correlations of PC1 with activation norm and token frequency in Pythia-1.4B.}{tab:app-pc1-associations}{Tokens included & Count & $|r|$: norm & 95\% interval & $\rho$: frequency & 95\% interval}
All & 65,535 & 0.999808 & [0.999786, 0.999826] & 0.543 & [0.532, 0.553] \\
First position & 862 & 0.999994 & [0.999993, 0.999994] & -0.621 & [-0.684, -0.568] \\
Other positions & 64,673 & 0.999616 & [0.999568, 0.999656] & 0.552 & [0.540, 0.562] \\
\end{apptable}

\subsection{Predicting model quantities}

We fit linear ridge regressions on calibration articles and test on separate articles (Table~\ref{tab:app-pc1-prediction}). The targets are next-token entropy and surprisal, the current token's log training frequency, log token position, and attention to the first token in the next layer, averaged across heads. The first position is excluded from the attention prediction.

For each regression, we standardize the inputs and choose the ridge penalty with five-fold cross-validation. We try 15 log-spaced penalties from $10^{-3}$ to $10^4$ for a scalar input, or 19 from $10^{-3}$ to $10^6$ for a full vector. Scalar results average five split seeds, and vector results average two.

Table~\ref{tab:app-pc1-prediction} gives test $R^2$ for PC1, norm, and full activations with or without PC1 (Raw and $Q\mathbf x$). Each Random column is the 95th percentile over 32 random unit directions; Random $\perp$ restricts these to directions orthogonal to PC1. We also fit regressions using PC2 and log position.

\begin{apptable}{llrrrrrr}{Test $R^2$ for predicting model quantities.}{tab:app-pc1-prediction}{Model & Target & PC1 & Norm & Random & Random $\perp$ & $Q\mathbf x$ & Raw}
GPT-2 & Entropy & 0.057 & 0.052 & 0.062 & 0.013 & 0.597 & 0.649 \\
GPT-2 & Surprisal & 0.019 & 0.017 & 0.023 & 0.005 & 0.241 & 0.258 \\
GPT-2 & Frequency & 0.000 & 0.000 & 0.012 & 0.062 & 0.938 & 0.938 \\
GPT-2 & Position & 0.170 & 0.169 & 0.207 & 0.078 & 0.826 & 0.988 \\
GPT-2 & Sink attention & 0.000 & 0.058 & 0.040 & 0.039 & 0.896 & 0.896 \\
Pythia-160M & Entropy & 0.040 & 0.029 & 0.064 & 0.046 & 0.687 & 0.697 \\
Pythia-160M & Surprisal & 0.024 & 0.020 & 0.028 & 0.010 & 0.198 & 0.204 \\
Pythia-160M & Frequency & 0.002 & 0.001 & 0.048 & 0.108 & 0.915 & 0.916 \\
Pythia-160M & Position & 0.185 & 0.172 & 0.185 & 0.056 & 0.882 & 0.915 \\
Pythia-160M & Sink attention & -0.004 & -0.004 & 0.036 & 0.060 & 0.735 & 0.736 \\
Pythia-1.4B & Entropy & 0.048 & 0.044 & 0.062 & 0.035 & 0.736 & 0.736 \\
Pythia-1.4B & Surprisal & 0.046 & 0.043 & 0.053 & 0.015 & 0.323 & 0.324 \\
Pythia-1.4B & Frequency & 0.002 & 0.001 & 0.048 & 0.088 & 0.926 & 0.926 \\
Pythia-1.4B & Position & 0.205 & 0.201 & 0.207 & 0.050 & 0.925 & 0.926 \\
Pythia-1.4B & Sink attention & 0.011 & 0.010 & 0.031 & 0.047 & 0.850 & 0.850 \\
Pythia-6.9B & Entropy & 0.058 & 0.055 & 0.063 & 0.016 & 0.761 & 0.762 \\
Pythia-6.9B & Surprisal & 0.054 & 0.052 & 0.056 & 0.007 & 0.341 & 0.342 \\
Pythia-6.9B & Frequency & 0.002 & 0.001 & 0.013 & 0.025 & 0.923 & 0.923 \\
Pythia-6.9B & Position & 0.208 & 0.206 & 0.209 & 0.022 & 0.931 & 0.936 \\
Pythia-6.9B & Sink attention & 0.016 & 0.016 & 0.032 & 0.023 & 0.744 & 0.743 \\
Llama-3.1-8B & Entropy & 0.024 & 0.024 & 0.035 & 0.034 & 0.780 & 0.624 \\
Llama-3.1-8B & Surprisal & 0.319 & 0.318 & 0.323 & 0.008 & 0.192 & 0.551 \\
Llama-3.1-8B & Frequency & 0.014 & 0.013 & 0.033 & 0.023 & 0.919 & 0.933 \\
Llama-3.1-8B & Position & 0.167 & 0.166 & 0.185 & 0.036 & 0.739 & 0.949 \\
Llama-3.1-8B & Sink attention & 0.009 & 0.166 & 0.068 & 0.068 & 0.942 & 0.943 \\
Qwen-2.5-7B & Entropy & 0.043 & 0.042 & 0.045 & 0.021 & 0.677 & 0.673 \\
Qwen-2.5-7B & Surprisal & 0.061 & 0.060 & 0.062 & 0.013 & 0.308 & 0.308 \\
Qwen-2.5-7B & Frequency & 0.000 & 0.000 & 0.001 & 0.048 & 0.916 & 0.915 \\
Qwen-2.5-7B & Position & 0.189 & 0.190 & 0.193 & 0.062 & 0.948 & 0.959 \\
Qwen-2.5-7B & Sink attention & -0.000 & -0.000 & 0.001 & 0.009 & 0.214 & 0.193 \\
Gemma-2-2B & Entropy & 0.021 & 0.020 & 0.040 & 0.026 & 0.681 & 0.697 \\
Gemma-2-2B & Surprisal & 0.153 & 0.153 & 0.158 & 0.016 & 0.235 & 0.378 \\
Gemma-2-2B & Frequency & 0.005 & 0.002 & 0.051 & 0.081 & 0.940 & 0.945 \\
Gemma-2-2B & Position & 0.191 & 0.196 & 0.226 & 0.071 & 0.789 & 0.964 \\
Gemma-2-2B & Sink attention & 0.011 & 0.009 & 0.062 & 0.062 & 0.898 & 0.898 \\
Gemma-2-9B & Entropy & 0.044 & 0.042 & 0.045 & 0.012 & 0.663 & 0.698 \\
Gemma-2-9B & Surprisal & 0.185 & 0.181 & 0.145 & 0.006 & 0.200 & 0.370 \\
Gemma-2-9B & Frequency & 0.003 & 0.000 & 0.046 & 0.050 & 0.938 & 0.942 \\
Gemma-2-9B & Position & 0.258 & 0.256 & 0.214 & 0.041 & 0.675 & 0.967 \\
Gemma-2-9B & Sink attention & 0.126 & 0.123 & 0.161 & 0.131 & 0.913 & 0.913 \\
\end{apptable}

\subsection{Removing and restoring PC1}
\label{app:pc1_interventions}

We replace the PC1 coefficient with its calibration mean, $\mathbf h\mapsto\mathbf h-P(\mathbf h-\boldsymbol\mu)$, or with its mean at the same token position, and compare these with removing PC2 or a random direction (Table~\ref{tab:app-pc1-delete}); random results average five directions.

\begin{apptable}{lrrrrr}{Cross-entropy after residual-stream interventions, in nats per prediction.}{tab:app-pc1-delete}{Model & Unedited & Delete PC1 & Position mean & Delete PC2 & Random}
GPT-2 & 3.929 & 4.728 & 3.933 & 4.219 & 3.938 \\
Pythia-160M & 3.975 & 4.198 & 4.054 & 3.997 & 3.986 \\
Pythia-1.4B & 3.149 & 3.694 & 3.292 & 3.174 & 3.162 \\
Pythia-6.9B & 2.968 & 3.379 & 3.271 & 3.046 & 2.969 \\
Llama-3.1-8B & 2.600 & 4.610 & 2.600 & 2.747 & 2.599 \\
Qwen-2.5-7B & 2.578 & 8.443 & 4.437 & 2.575 & 2.585 \\
Gemma-2-2B & 3.004 & 3.188 & 3.007 & 2.981 & 3.005 \\
Gemma-2-9B & 2.759 & 3.089 & 2.761 & 2.665 & 2.759 \\
\end{apptable}

Table~\ref{tab:app-writers} gives loss after removing PC1 from an earlier MLP output ($\mathbf m-P\mathbf m$). We choose the two earlier MLPs with the largest mean absolute output along the SAE layer's PC1. Random removes a random direction orthogonal to PC1. Restore later reinstates the original PC1 coefficient at the SAE layer. We also test edits confined to first or later positions and restoration immediately after removal.

\begin{apptable}{lrrrrr}{Cross-entropy after removing PC1 from MLP outputs, in nats per prediction.}{tab:app-writers}{Model & Layer & Unedited & Remove PC1 & Random $\perp$ & Restore later}
GPT-2 & 2 & 3.929 & 3.913 & 3.930 & 3.927 \\
GPT-2 & 0 & 3.929 & 3.926 & 3.929 & 3.927 \\
Pythia-160M & 3 & 3.975 & 4.746 & 3.976 & 4.520 \\
Pythia-160M & 0 & 3.975 & 4.105 & 3.977 & 4.101 \\
Pythia-1.4B & 3 & 3.149 & 3.806 & 3.149 & 3.808 \\
Pythia-1.4B & 4 & 3.149 & 3.176 & 3.149 & 3.146 \\
Pythia-6.9B & 4 & 2.969 & 3.209 & 2.968 & 3.028 \\
Pythia-6.9B & 5 & 2.969 & 3.121 & 2.970 & 2.980 \\
Llama-3.1-8B & 1 & 2.600 & 10.910 & 2.600 & 10.819 \\
Llama-3.1-8B & 10 & 2.600 & 2.600 & 2.600 & 2.600 \\
Qwen-2.5-7B & 3 & 2.579 & 10.111 & 2.580 & 10.381 \\
Qwen-2.5-7B & 4 & 2.579 & 2.703 & 2.580 & 2.581 \\
Gemma-2-2B & 7 & 3.004 & 2.999 & 3.004 & 3.004 \\
Gemma-2-2B & 11 & 3.004 & 2.994 & 3.004 & 3.002 \\
Gemma-2-9B & 0 & 2.758 & 2.765 & 2.757 & 2.784 \\
Gemma-2-9B & 16 & 2.758 & 2.746 & 2.759 & 2.761 \\
\end{apptable}

\section{Layers, Positions, and Principal-Component Controls}
\label{app:multilayer}

\subsection{Variance across layers}

Table~\ref{tab:app-layer-profile} reports variance explained by PC1 and PC2 at each Pythia-160M layer, including the output before and after final normalization. We use 65,705 activations from 88 articles, computing model outputs in float32 and centered covariance in float64. Untrained PC1 is the range across three randomly initialized models, with seeds 0, 1, and 2, evaluated on the same inputs.

\begin{apptable}{lrrr}{Variance explained (\%) across Pythia-160M layers.}{tab:app-layer-profile}{Activations & Trained PC1 & Trained PC2 & Untrained PC1}
Embedding & 5.02 & 4.42 & 4.49--4.88 \\
Layer 0 & 9.82 & 4.24 & 4.35--4.64 \\
Layer 1 & 21.02 & 4.66 & 4.23--4.89 \\
Layer 2 & 59.39 & 2.53 & 3.90--5.04 \\
Layer 3 & 92.70 & 0.81 & 3.84--4.93 \\
Layer 4 & 93.28 & 0.74 & 3.81--4.55 \\
Layer 5 & 92.80 & 0.79 & 3.73--4.39 \\
Layer 6 & 92.40 & 0.69 & 3.82--4.42 \\
Layer 7 & 90.75 & 0.90 & 4.06--4.41 \\
Layer 8 & 87.33 & 1.15 & 4.11--4.52 \\
Layer 9 & 78.05 & 2.03 & 4.06--4.57 \\
Layer 10 & 45.39 & 7.97 & 4.01--4.53 \\
Layer 11 & 64.01 & 1.90 & 4.05--4.41 \\
Final normalization & 5.97 & 2.19 & 3.88--4.07 \\
\end{apptable}

\subsection{SAEs at additional layers}

Table~\ref{tab:app-layer-extension} gives standard/bypass sharing and sparsity at two additional layers per model. We use five seeds and the default training settings, with 1,500 features ranked on 64 validation sequences. Sharing averages all 20 ordered seed pairs.

\begin{apptable}{lrrrr}{Feature sharing (\%) and sparsity at additional layers.}{tab:app-layer-extension}{Model & Layer & Top 50 & Dictionary & $L_0$}
GPT-2 & 0 & 53.7/50.7 & 6.2/6.0 & 63.94/64.00 \\
GPT-2 & 11 & 94.0/95.1 & 58.4/58.5 & 64.00/64.00 \\
Pythia-160M & 1 & 48.6/50.0 & 7.4/7.3 & 64.00/64.00 \\
Pythia-160M & 9 & 18.4/93.1 & 54.7/53.4 & 63.98/64.00 \\
Pythia-1.4B & 2 & 55.8/54.9 & 15.2/14.6 & 63.98/64.00 \\
Pythia-1.4B & 23 & 87.0/90.8 & 59.3/59.2 & 63.90/63.98 \\
Pythia-6.9B & 2 & 58.4/57.9 & 20.2/17.5 & 63.84/64.00 \\
Pythia-6.9B & 31 & 91.8/92.6 & 60.1/58.8 & 63.65/63.77 \\
Llama-3.1-8B & 7 & 67.8/96.4 & 64.5/65.1 & 63.68/63.27 \\
Llama-3.1-8B & 31 & 92.1/93.4 & 64.5/65.2 & 63.38/63.81 \\
Qwen-2.5-7B & 2 & 78.0/79.1 & 36.1/34.1 & 63.85/64.00 \\
Qwen-2.5-7B & 27 & 90.4/95.3 & 68.3/68.5 & 63.76/64.00 \\
Gemma-2-2B & 2 & 58.3/57.4 & 16.0/15.8 & 64.00/63.99 \\
Gemma-2-2B & 22 & 92.3/94.1 & 67.5/70.4 & 64.00/63.60 \\
Gemma-2-9B & 9 & 94.0/95.0 & 70.3/73.8 & 63.96/63.39 \\
Gemma-2-9B & 33 & 83.4/94.9 & 70.7/74.3 & 64.00/63.76 \\
\end{apptable}

Table~\ref{tab:app-layers} gives top-50 sharing ($S_{50}$), dictionary sharing, and loss with SAE reconstructions (CE) for a separate Pythia-1.4B experiment. We use layers 3, 12, and 23, before final normalization. Here the data are \texttt{wikitext-103-raw-v1} entries joined with newlines and split into 128-token sequences. We use seeds 42 and 137 and the default 8,192-feature training settings (Table~\ref{tab:app-training}). Model outputs are float32. PCA is fitted on a random sample of 50,000 training activations; feature rankings use 64 validation sequences.

\begin{apptable}{lrrrrrrr}{Pythia-1.4B feature sharing (\%) and cross-entropy across layers.}{tab:app-layers}{Layer & $w_1$ (\%) & $S_{50}$ S & $S_{50}$ B & Dict. S & Dict. B & CE S & CE B}
3 & 93.82 & 14.0 & 68.0 & 30.0 & 23.0 & 3.25249 & 3.24169 \\
12 & 92.95 & 8.0 & 92.0 & 63.2 & 62.4 & 3.22981 & 3.22407 \\
23 & 11.09 & 96.0 & 92.0 & 64.6 & 63.7 & 3.67123 & 3.66968 \\
\end{apptable}

\subsection{Removing additional principal components}
\label{app:pcsweep}

Table~\ref{tab:app-pcsweep} gives top-50 sharing after removing leading PCs from Pythia-160M layer-6 activations. We use the settings in Table~\ref{tab:app-training} with 4,096 features, one million training tokens, float32 model outputs, and seeds 42 and 137. The first 50,000 activations are used for PCA; 64 validation sequences are used for sharing.

\begin{apptable}{lrrrrrrr}{Pythia-160M top-50 sharing after removing leading principal components.}{tab:app-pcsweep}{PCs removed & 0 & 1 & 2 & 5 & 10 & 20 & 50}
Sharing (\%) & 4 & 86 & 90 & 84 & 80 & 74 & 84 \\
\end{apptable}

\subsection{Excluding first positions and special tokens}
\label{app:bos}

Table~\ref{tab:app-positions} gives PC1 variance after refitting on different sets of tokens, from up to 128 validation sequences. After first omits array position zero; No special also omits tokenizer special tokens. Counts exclude padding and are taken before the other exclusions. No added BOS is a separate model run with automatic special-token addition disabled.

\begin{apptable}{lrllrrr}{PC1 variance share (\%) with different token exclusions and tokenization settings.}{tab:app-positions}{Model & Context & Tokenization & Tokens & All & After first & No special}
GPT-2 & 128 & Default & 9811 & 96.66 & 3.89 & 3.89 \\
GPT-2 & 512 & Default & 12831 & 95.56 & 3.94 & 3.94 \\
Pythia-160M & 128 & Default & 9821 & 92.41 & 88.16 & 88.16 \\
Pythia-160M & 512 & Default & 13103 & 91.06 & 87.06 & 87.06 \\
Pythia-1.4B & 128 & Default & 9821 & 95.65 & 91.60 & 91.60 \\
Pythia-1.4B & 512 & Default & 13103 & 95.14 & 92.01 & 92.01 \\
Llama-3.1-8B & 128 & Default & 10015 & 96.59 & 2.85 & 2.85 \\
Llama-3.1-8B & 128 & No added BOS & 9936 & 98.24 & 91.25 & 91.25 \\
Llama-3.1-8B & 512 & Default & 13597 & 95.45 & 2.64 & 2.64 \\
Llama-3.1-8B & 512 & No added BOS & 13469 & 97.65 & 88.52 & 88.52 \\
Qwen-2.5-7B & 128 & Default & 10093 & 99.87 & 99.68 & 99.68 \\
Qwen-2.5-7B & 512 & Default & 14160 & 99.83 & 99.56 & 99.56 \\
Gemma-2-2B & 128 & Default & 10111 & 76.52 & 66.73 & 14.87 \\
Gemma-2-2B & 128 & No added BOS & 10033 & 8.79 & 8.69 & 8.72 \\
Gemma-2-2B & 512 & Default & 13810 & 70.79 & 70.79 & 11.84 \\
Gemma-2-2B & 512 & No added BOS & 13682 & 8.45 & 8.45 & 8.48 \\
\end{apptable}

\section{Released SAEs and Preprocessing}

\subsection{Released SAEs}
\label{app:released}

We evaluate released SAEs on WikiText-103, using 50,000 activations for calibration and up to 100,000 further activations for evaluation, with 128-token contexts.

Gemma Scope, Gemma Scope 2, and Llama Scope \citep{lieberum2024gemmascope,mcdougall2025gemmascope2,he2024llamascope} exclude start, end, and padding tokens. We use their released weights directly. For OpenAI GPT-2 \citep{gao2024scaling}, we subtract the coordinate mean, normalize the vector, and subtract decoder bias before encoding; FVU uses the normalized inputs.

Table~\ref{tab:app-released} gives total and content FVU for 44 checkpoints. Layer and Tag are the release labels. $L_0$ is measured on our data; $1\mathrm{k}=1024$ and $R=\mathrm{FVU}_{\perp}/\mathrm{FVU}_{\rm total}$.

\begin{apptable}{lrrlrrrr}{Total and content FVU for released SAEs.}{tab:app-released}{Model & Layer & Width & Tag & $L_0$ & Total FVU & Content FVU & $R$}
\multicolumn{8}{l}{\textit{Gemma Scope}}\\
Gemma-2-2B & 6 & 16k & 36 & 38.3 & 0.22550 & 0.23807 & 1.056 \\
Gemma-2-2B & 6 & 16k & 70 & 76.1 & 0.18626 & 0.19670 & 1.056 \\
Gemma-2-2B & 6 & 64k & 107 & 118.4 & 0.12121 & 0.12802 & 1.056 \\
Gemma-2-2B & 6 & 64k & 56 & 60.9 & 0.15026 & 0.15866 & 1.056 \\
Gemma-2-2B & 12 & 16k & 41 & 42.0 & 0.24474 & 0.30489 & 1.246 \\
Gemma-2-2B & 12 & 16k & 82 & 81.5 & 0.20612 & 0.25682 & 1.246 \\
Gemma-2-2B & 12 & 64k & 38 & 38.0 & 0.21258 & 0.26483 & 1.246 \\
Gemma-2-2B & 12 & 64k & 72 & 73.6 & 0.17669 & 0.22015 & 1.246 \\
Gemma-2-2B & 22 & 16k & 38 & 38.4 & 0.28973 & 0.30137 & 1.040 \\
Gemma-2-2B & 22 & 16k & 72 & 72.4 & 0.24844 & 0.25852 & 1.041 \\
Gemma-2-2B & 22 & 64k & 116 & 121.2 & 0.17346 & 0.18051 & 1.041 \\
Gemma-2-2B & 22 & 64k & 62 & 63.6 & 0.20709 & 0.21548 & 1.041 \\
Gemma-2-9B & 9 & 16k & 100 & 113.9 & 0.22771 & 0.23644 & 1.038 \\
Gemma-2-9B & 9 & 16k & 51 & 57.7 & 0.26426 & 0.27425 & 1.038 \\
Gemma-2-9B & 9 & 128k & 42 & 49.4 & 0.17345 & 0.18003 & 1.038 \\
Gemma-2-9B & 9 & 128k & 75 & 88.8 & 0.14682 & 0.15246 & 1.038 \\
Gemma-2-9B & 21 & 16k & 129 & 127.2 & 0.18127 & 0.28192 & 1.555 \\
Gemma-2-9B & 21 & 16k & 36 & 35.1 & 0.24853 & 0.38642 & 1.555 \\
Gemma-2-9B & 21 & 128k & 109 & 110.0 & 0.14596 & 0.22699 & 1.555 \\
Gemma-2-9B & 21 & 128k & 58 & 57.8 & 0.17516 & 0.27237 & 1.555 \\
Gemma-2-9B & 35 & 16k & 120 & 121.9 & 0.25059 & 0.26311 & 1.050 \\
Gemma-2-9B & 35 & 16k & 61 & 60.5 & 0.29364 & 0.30825 & 1.050 \\
Gemma-2-9B & 35 & 128k & 51 & 51.1 & 0.22114 & 0.23212 & 1.050 \\
Gemma-2-9B & 35 & 128k & 94 & 95.2 & 0.18734 & 0.19668 & 1.050 \\
\multicolumn{8}{l}{\textit{Gemma Scope 2}}\\
Gemma-3-1B & 7 & 16k & medium & 58.2 & 0.06124 & 0.12955 & 2.116 \\
Gemma-3-1B & 7 & 16k & small & 20.2 & 0.07998 & 0.16794 & 2.100 \\
Gemma-3-1B & 7 & 64k & medium & 57.3 & 0.05039 & 0.10590 & 2.102 \\
Gemma-3-1B & 7 & 64k & small & 20.1 & 0.06831 & 0.14159 & 2.073 \\
Gemma-3-1B & 13 & 16k & medium & 61.2 & 0.12413 & 0.27350 & 2.203 \\
Gemma-3-1B & 13 & 16k & small & 21.5 & 0.16630 & 0.36211 & 2.178 \\
Gemma-3-1B & 13 & 64k & medium & 60.4 & 0.11313 & 0.24833 & 2.195 \\
Gemma-3-1B & 13 & 64k & small & 21.2 & 0.15443 & 0.33454 & 2.166 \\
Gemma-3-1B & 22 & 16k & medium & 61.9 & 0.20049 & 0.27266 & 1.360 \\
Gemma-3-1B & 22 & 16k & small & 23.6 & 0.25709 & 0.34737 & 1.351 \\
Gemma-3-1B & 22 & 64k & medium & 61.6 & 0.17096 & 0.23173 & 1.355 \\
Gemma-3-1B & 22 & 64k & small & 23.5 & 0.22078 & 0.29734 & 1.347 \\
\multicolumn{8}{l}{\textit{Llama Scope}}\\
Llama-3.1-8B & 6 & 32k & -- & 31.0 & 0.40442 & 0.41059 & 1.015 \\
Llama-3.1-8B & 6 & 128k & -- & 35.3 & 0.33362 & 0.33949 & 1.018 \\
Llama-3.1-8B & 16 & 32k & -- & 32.0 & 0.37068 & 0.37834 & 1.021 \\
Llama-3.1-8B & 16 & 128k & -- & 35.1 & 0.31843 & 0.32480 & 1.020 \\
Llama-3.1-8B & 27 & 32k & -- & 29.4 & 0.41986 & 0.44304 & 1.055 \\
Llama-3.1-8B & 27 & 128k & -- & 32.2 & 0.35982 & 0.37968 & 1.055 \\
\multicolumn{8}{l}{\textit{OpenAI GPT-2}}\\
GPT-2 & 8 & 32k & -- & 32.0 & 0.17666 & 0.18238 & 1.032 \\
GPT-2 & 8 & 128k & -- & 32.0 & 0.14626 & 0.15093 & 1.032 \\
\end{apptable}

\paragraph{Qwen Scope.}
Table~\ref{tab:app-qwen-release} gives reconstruction results for six Qwen Scope SAEs on Qwen3-8B-Base \citep{deng2026qwenscope}. Qwen Scope filters large-norm activations, but the exact threshold is not specified. We therefore try 5, 10, and 20 times the median raw norm in the calibration data, filtering both calibration and test activations. All SAEs have 65,536 features; $k$ is the release's sparsity label, and Removed is the percentage of evaluation activations excluded.

\begin{apptable}{rrlrrrrr}{Qwen Scope reconstruction after excluding high-norm activations.}{tab:app-qwen-release}{Layer & $k$ & Cutoff & Removed (\%) & $w_1$ (\%) & Total FVU & Content FVU & $R$}
7 & 50 & All & 0.00 & 99.71 & 1.02107 & 178.42778 & 174.746 \\
7 & 50 & 5$\times$ & 0.80 & 2.91 & 0.24005 & 0.24709 & 1.029 \\
7 & 50 & 10$\times$ & 0.80 & 2.91 & 0.24005 & 0.24709 & 1.029 \\
7 & 50 & 20$\times$ & 0.80 & 2.91 & 0.24005 & 0.24709 & 1.029 \\
7 & 100 & All & 0.00 & 99.71 & 1.10307 & 242.81489 & 220.125 \\
7 & 100 & 5$\times$ & 0.80 & 2.91 & 0.20789 & 0.21402 & 1.029 \\
7 & 100 & 10$\times$ & 0.80 & 2.91 & 0.20789 & 0.21402 & 1.029 \\
7 & 100 & 20$\times$ & 0.80 & 2.91 & 0.20789 & 0.21402 & 1.029 \\
18 & 50 & All & 0.00 & 99.10 & 0.22524 & 20.76911 & 92.209 \\
18 & 50 & 5$\times$ & 0.82 & 2.54 & 0.32881 & 0.33690 & 1.025 \\
18 & 50 & 10$\times$ & 0.82 & 2.54 & 0.32881 & 0.33690 & 1.025 \\
18 & 50 & 20$\times$ & 0.82 & 2.54 & 0.32881 & 0.33690 & 1.025 \\
18 & 100 & All & 0.00 & 99.10 & 0.14942 & 15.73571 & 105.313 \\
18 & 100 & 5$\times$ & 0.82 & 2.54 & 0.27656 & 0.28349 & 1.025 \\
18 & 100 & 10$\times$ & 0.82 & 2.54 & 0.27656 & 0.28349 & 1.025 \\
18 & 100 & 20$\times$ & 0.82 & 2.54 & 0.27656 & 0.28349 & 1.025 \\
30 & 50 & All & 0.00 & 74.97 & 0.07401 & 0.29520 & 3.988 \\
30 & 50 & 5$\times$ & 0.82 & 7.91 & 0.28395 & 0.30821 & 1.085 \\
30 & 50 & 10$\times$ & 0.82 & 7.99 & 0.28308 & 0.30754 & 1.086 \\
30 & 50 & 20$\times$ & 0.00 & 74.94 & 0.07414 & 0.29527 & 3.982 \\
30 & 100 & All & 0.00 & 74.97 & 0.06269 & 0.25042 & 3.994 \\
30 & 100 & 5$\times$ & 0.82 & 7.91 & 0.24174 & 0.26243 & 1.086 \\
30 & 100 & 10$\times$ & 0.82 & 7.99 & 0.24100 & 0.26186 & 1.087 \\
30 & 100 & 20$\times$ & 0.00 & 74.94 & 0.06280 & 0.25048 & 3.988 \\
\end{apptable}

\paragraph{Pythia and context length.}
Table~\ref{tab:app-pythia-context} evaluates the released Pythia-160M layer-6 SAE, which has 32,768 features and $k=32$ and subtracts decoder bias before encoding. We use separate 128- or 512-token chunks without added BOS, or 2,049-token contexts formed by joining text with end-of-sequence (EOS) separators. After first excludes the first position from evaluation; Exclude EOS omits the separators.

\begin{apptable}{rlrrrrr}{Released Pythia-160M SAE reconstruction at different context lengths.}{tab:app-pythia-context}{Context & Tokens included & Count & $w_1$ (\%) & Total FVU & Content FVU & $R$}
128 & After first & 99,211 & 80.04 & 0.04049 & 0.20275 & 5.007 \\
128 & All & 100,000 & 87.61 & 0.02407 & 0.19416 & 8.067 \\
512 & After first & 99,795 & 54.99 & 0.08597 & 0.19084 & 2.220 \\
512 & All & 100,000 & 66.63 & 0.06303 & 0.18873 & 2.994 \\
2049 & All packed & 100,000 & 33.03 & 0.11928 & 0.17796 & 1.492 \\
2049 & Exclude EOS & 99,978 & 31.59 & 0.12201 & 0.17819 & 1.460 \\
\end{apptable}

\subsection{Normalization, filtering, and whitening}
\label{app:preproc}

We compare preprocessing methods with the default training settings and seeds 42, 137, and 0. We fit preprocessing parameters on the first 200,000 training activations and evaluate on 64 validation sequences. Before measuring reconstruction error or model loss, we undo the preprocessing.

Unit-norm preprocessing divides each centered activation by its norm and saves the norm to rescale the reconstruction. Coordinate normalization also subtracts the mean across coordinates, saving both that mean and the norm. Norm filtering removes training activations whose centered norm exceeds ten times the median training norm.

Whitening uses the training covariance $V\operatorname{diag}(\lambda_i)V^\top$:
\[
 \mathbf y=(\mathbf x-\boldsymbol\mu)V
 \operatorname{diag}((\lambda_i+\epsilon)^{-1/2}),
 \qquad \epsilon=10^{-4}\lambda_1.
\]
We invert this transform after reconstruction.

Tables~\ref{tab:app-preproc} and~\ref{tab:app-retained} give reconstruction results on all tokens and on tokens below the norm cutoff, respectively, averaged over three seeds. Coefficients counts sparse features plus the extra values needed to undo preprocessing. CE is loss with SAE reconstructions.

\begin{apptable}{lllrrrr}{Reconstruction after preprocessing, evaluated on all tokens.}{tab:app-preproc}{Model & Method & Coefficients & $L_0$ & Total FVU & Content FVU & CE}
GPT-2 & Standard & 64+0 & 64.00 & 0.004348 & 0.13650 & 4.11441 \\
GPT-2 & Bypass 63+1 & 63+1 & 62.77 & 0.004276 & 0.13427 & 4.11617 \\
GPT-2 & Bypass 64+1 & 64+1 & 63.77 & 0.004243 & 0.13323 & 4.11435 \\
GPT-2 & Unit norm & 64+1 & 63.98 & 0.004398 & 0.13799 & 4.12948 \\
GPT-2 & Coord. norm & 64+2 & 64.00 & 0.004405 & 0.13822 & 4.12936 \\
GPT-2 & Coord. norm+bypass & 63+3 & 62.51 & 0.004665 & 0.14650 & 4.16077 \\
GPT-2 & Norm-filter train & 64+0 & 64.00 & 3.533275 & 76.87259 & 5.31978 \\
GPT-2 & Whiten & 64+0 & 64.00 & 0.004921 & 0.13433 & 4.13626 \\
Pythia-1.4B & Standard & 64+0 & 63.98 & 0.009074 & 0.22139 & 3.19588 \\
Pythia-1.4B & Bypass 63+1 & 63+1 & 63.00 & 0.008965 & 0.21876 & 3.18367 \\
Pythia-1.4B & Bypass 64+1 & 64+1 & 64.00 & 0.008936 & 0.21805 & 3.18461 \\
Pythia-1.4B & Unit norm & 64+1 & 64.00 & 0.009084 & 0.22153 & 3.19334 \\
Pythia-1.4B & Coord. norm & 64+2 & 64.00 & 0.009091 & 0.22169 & 3.20145 \\
Pythia-1.4B & Coord. norm+bypass & 63+3 & 61.82 & 0.009115 & 0.22240 & 3.21050 \\
Pythia-1.4B & Norm-filter train & 64+0 & 64.00 & 0.079522 & 1.85201 & 3.50372 \\
Pythia-1.4B & Whiten & 64+0 & 64.00 & 0.009515 & 0.21826 & 3.20243 \\
Qwen-2.5-7B & Standard & 64+0 & 63.98 & 0.000398 & 0.32329 & 2.50462 \\
Qwen-2.5-7B & Bypass 63+1 & 63+1 & 62.95 & 0.000248 & 0.20184 & 2.46515 \\
Qwen-2.5-7B & Bypass 64+1 & 64+1 & 63.95 & 0.000247 & 0.20109 & 2.46161 \\
Qwen-2.5-7B & Coord. norm & 64+2 & 63.93 & 0.000586 & 0.47437 & 2.65635 \\
Qwen-2.5-7B & Norm-filter train & 64+0 & 63.97 & 0.315440 & 253.28876 & 2.86624 \\
Qwen-2.5-7B & Whiten & 64+0 & 63.99 & 0.000962 & 0.19604 & 2.46488 \\
Gemma-2-2B & Standard & 64+0 & 63.90 & 0.043965 & 0.19754 & 3.04176 \\
Gemma-2-2B & Bypass 63+1 & 63+1 & 62.82 & 0.043429 & 0.19522 & 3.03572 \\
Gemma-2-2B & Bypass 64+1 & 64+1 & 63.77 & 0.043272 & 0.19451 & 3.03512 \\
Gemma-2-2B & Coord. norm & 64+2 & 63.82 & 0.043401 & 0.19503 & 3.03799 \\
Gemma-2-2B & Norm-filter train & 64+0 & 63.90 & 6.437689 & 24.27273 & 3.76865 \\
Gemma-2-2B & Whiten & 64+0 & 64.00 & 0.049051 & 0.19986 & 3.04060 \\
\end{apptable}

\begin{apptable}{llrrr}{Reconstruction evaluated on tokens below the norm cutoff.}{tab:app-retained}{Model & Method & Tokens & Total FVU & Content FVU}
GPT-2 & Standard & 4545 & 0.094406 & 0.13591 \\
GPT-2 & Bypass 63+1 & 4545 & 0.093174 & 0.13416 \\
GPT-2 & Norm-filter train & 4545 & 0.092957 & 0.13384 \\
GPT-2 & Whiten & 4545 & 0.106670 & 0.13429 \\
Pythia-1.4B & Standard & 4461 & 0.128739 & 0.22867 \\
Pythia-1.4B & Bypass 63+1 & 4461 & 0.127931 & 0.22725 \\
Pythia-1.4B & Norm-filter train & 4461 & 0.128220 & 0.22774 \\
Pythia-1.4B & Whiten & 4461 & 0.132677 & 0.22671 \\
Qwen-2.5-7B & Standard & 4593 & 0.013540 & 0.27829 \\
Qwen-2.5-7B & Bypass 63+1 & 4593 & 0.011463 & 0.23562 \\
Qwen-2.5-7B & Norm-filter train & 4593 & 0.016704 & 0.34329 \\
Qwen-2.5-7B & Whiten & 4593 & 0.011431 & 0.22822 \\
Gemma-2-2B & Standard & 4624 & 0.173933 & 0.19784 \\
Gemma-2-2B & Bypass 63+1 & 4624 & 0.171828 & 0.19553 \\
Gemma-2-2B & Norm-filter train & 4624 & 0.171564 & 0.19516 \\
Gemma-2-2B & Whiten & 4624 & 0.194046 & 0.20018 \\
\end{apptable}

\section{Vision Transformers}
\label{app:vision}

\subsection{Models and SAE training}

We use DINOv2-base and DINOv2-with-registers-base \citep{oquab2023dinov2,darcet2024vision}, with 512 training and 128 test images drawn from CIFAR-100's training split. We take the $16\times16$ patch activations after layer 8, before final normalization, omitting CLS and register tokens.

We train BatchTopK SAEs with 6,144 features and mean $L_0=32$, using the settings in Table~\ref{tab:app-training} for 10,000 steps. Here we normalize decoder vectors only at initialization. The BatchTopK cutoff is recalculated for each batch, including test batches, and tied values are retained. We use float32 with seeds 0, 1, and 2. Random bypass uses seed 4242. Both bypasses have one extra coefficient.

\subsection{Registers}

Following \citet{jiang2025vision}, we redirect two or 50 MLP neurons to an added register token. We run 32 training images through the model, keep patches with norms above ten times the median, and rank the neurons in layers 0--8 by their activity on these patches. The register starts at zero. At each layer we find the largest activation in absolute value among the selected neurons over the original token positions, and copy it, including its sign, to the selected register coordinates. The pretrained comparison uses DINOv2 with four registers.

\subsection{Reconstruction and feature matching}

Table~\ref{tab:app-vision-fidelity} reports reconstruction for each model and register setting. FVU is the mean $\pm$ standard deviation across SAE seeds. PC1 variance $w_1$ and the maximum-to-median norm ratio are measured on training patches. $R$ is mean content FVU divided by mean total FVU.

\begin{apptable}{lrrrrr}{PC1 variance, activation norms, and reconstruction in vision models.}{tab:app-vision-fidelity}{Condition & $w_1$ (\%) & Max/median & Total FVU & Content FVU & $R$}
Standard & 96.19 & 57.24 & $0.01310\pm0.00009$ & $0.34349\pm0.00236$ & 26.21 \\
Random bypass & 96.19 & 57.24 & $0.01312\pm0.00009$ & $0.34392\pm0.00224$ & 26.21 \\
PC1 bypass & 96.19 & 57.24 & $0.01268\pm0.00004$ & $0.33277\pm0.00096$ & 26.24 \\
Test-time, 50 neurons & 3.96 & 2.32 & $0.41473\pm0.00172$ & $0.42879\pm0.00163$ & 1.03 \\
Test-time, 2 neurons & 75.66 & 18.57 & $0.08480\pm0.00029$ & $0.34931\pm0.00125$ & 4.12 \\
Pretrained registers & 4.00 & 1.71 & $0.42751\pm0.00035$ & $0.44135\pm0.00015$ & 1.03 \\
\end{apptable}

For Table~\ref{tab:app-vision-sharing}, we normalize decoder directions and match features across all six directed comparisons among three seeds. The table reports the mean of the per-comparison median best-match cosine and sharing rate over all active features.

\begin{apptable}{lrrrr}{Feature matching across vision SAEs.}{tab:app-vision-sharing}{Condition & $L_0$ & Alive & Median best cosine & Shared $>0.7$ (\%)}
Standard & 32.00 & 6,144 & 0.3954 & 28.84 \\
Random bypass & 32.00 & 6,144 & 0.3924 & 28.50 \\
PC1 bypass & 32.00 & 6,144 & 0.4186 & 16.62 \\
Test-time, 50 neurons & 32.00 & 6,144 & 0.4655 & 13.84 \\
Test-time, 2 neurons & 32.00 & 6,144 & 0.4531 & 25.28 \\
Pretrained registers & 32.00 & 6,144 & 0.4793 & 15.91 \\
\end{apptable}

\section{Crosscoders}
\label{app:crosscoders}

\subsection{Training and evaluation}

Crosscoder settings follow Table~\ref{tab:app-training}, with the following changes and additions:
\begin{center}
\begin{tabular}{p{0.30\linewidth}p{0.62\linewidth}}
\toprule
Setting & Value \\
\midrule
Features & 8,192 \\
Training tokens & Approximately three million \\
Initialization seeds & 0, 42, 137 \\
Data-order seed & 101 \\
Checkpoint selection & Lowest validation loss, checked every 500 steps \\
Mean, scale, and PCA sample & 50,000 training activations \\
Features ranked and matched & Features active on the test set \\
Matching & Signed decoder cosine $>0.7$, in original activation units \\
Sharing & Mean over all six directed seed pairs \\
FVU & Mean over three seeds, in original activation units \\
\bottomrule
\end{tabular}
\end{center}

We take at most 40 entries per article, truncating each to 128 tokens. Validation uses the ranking articles, and test uses the held-out articles in Table~\ref{tab:app-partitions}; neither enters training. Activations are stored in float16 and trained on in float32.

We subtract the input mean and divide by the root mean squared norm. Outputs use either a shared scale or separate scales (Table~\ref{tab:ext-crosscoder-arms}), with unit-length decoder vectors in the scaled coordinates. Bypass removes the chosen components from the decoder and bias after each training step.

For normalized target $\widetilde{\mathbf y}$ and reconstruction $\widehat{\widetilde{\mathbf y}}$, the importance of feature $j$ is
\[
 I_j^{\rm rec}=\mathbb E\!\left[
 \left\|\widehat{\widetilde{\mathbf y}}
       -z_j\mathbf d_j-\widetilde{\mathbf y}\right\|_2^2
 -\left\|\widehat{\widetilde{\mathbf y}}
       -\widetilde{\mathbf y}\right\|_2^2\right].
\]

\subsection{Activation pairs and reconstruction}

One encoder takes the two activation vectors concatenated, and one decoder reconstructs both. The four Pythia-1.4B pairs use the same tokens in both inputs:
\begin{enumerate}
\item Two copies of layer-12 activations.
\item The same copies, with the second copy's centered PC1 coefficient doubled.
\item Activations from layers 12 and 16.
\item Layer-12 activations at training step 71,000 and at the final checkpoint.
\end{enumerate}
For doubling, we fit the mean and PC1 on 200,000 training activations.

Table~\ref{tab:ext-crosscoder-arms} gives the methods. Table~\ref{tab:ext-crosscoder-sharing} gives sharing of the top 50 features ranked by $I_j^{\rm rec}$. Table~\ref{tab:ext-crosscoder-fidelity} gives dictionary sharing and content FVU, excluding each input's PC1.

\begin{apptable}{lp{0.78\textwidth}}{Crosscoder methods.}{tab:ext-crosscoder-arms}{Method & Definition}
X0 & Reconstruct both inputs with one output scale. \\
X1 & Reconstruct both inputs with a separate output scale for each. \\
X2 & Bypass PC1 in each input, saving two coefficients. \\
X3 & Bypass PC1 of the concatenated inputs, saving one coefficient. \\
X4 & Bypass one fixed random direction in each input, saving two coefficients. \\
\end{apptable}

\begin{apptable}{lrrrrr}{Cross-seed sharing (\%) of the top 50 crosscoder features.}{tab:ext-crosscoder-sharing}{Activation pair & X0 & X1 & X2 & X3 & X4}
Identical & 2.7 & 2.0 & 91.3 & 92.3 & 3.3 \\
PC1 doubled & 3.7 & 2.0 & 83.0 & 81.7 & 1.3 \\
Layers 12 and 16 & 2.7 & 3.3 & 93.0 & 92.7 & 3.3 \\
Step 71{,}000 and final & 2.7 & 2.0 & 93.0 & 92.7 & 1.3 \\
\end{apptable}

\begin{apptable}{lrrrr}{Dictionary sharing (\%) and content FVU for standard and PC1-bypass crosscoders.}{tab:ext-crosscoder-fidelity}{Activation pair & Dict. X0 & Dict. X2 & Content FVU X0 & Content FVU X2}
Identical & 66.8 & 67.2 & 0.3661 & 0.2753 \\
PC1 doubled & 60.4 & 66.6 & 0.4453 & 0.3053 \\
Layers 12 and 16 & 72.7 & 71.8 & 0.3495 & 0.2741 \\
Step 71{,}000 and final & 69.7 & 66.4 & 0.3561 & 0.2828 \\
\end{apptable}

\paragraph{Features exclusive to one input.}
Let A and B denote the first and second inputs in each pair. For decoder vectors in the original activation units, define
\[
 r_j=\frac{\|\mathbf d_{j,B}\|_2}
 {\|\mathbf d_{j,A}\|_2+\|\mathbf d_{j,B}\|_2}.
\]
A-only features have $r_j<0.1$; B-only features have $r_j>0.9$.

The two ReLU penalties, applied in the scaled coordinates, are
\[
 \underbrace{\lambda\mathbb E\!\left[\sum_jz_j\|\mathbf d_j\|_2\right]}_{\text{joint}}
 \qquad
 \underbrace{\lambda\mathbb E\!\left[
 \sum_jz_j\bigl(\|\mathbf d_{j,A}\|_2+
                    \|\mathbf d_{j,B}\|_2\bigr)\right]}_{\text{block-sum}}.
\]

Table~\ref{tab:ext-crosscoder-exclusive} gives sparsity and percentages of A-only and B-only features for seed 42 and $\lambda\in\{0.001,0.003,0.01\}$. The percentages are for block-sum. Joint-penalty and TopK runs produce no exclusive features. Both ReLU penalties also produce none on the identical-input and doubled-PC1 pairs.

\begin{apptable}{lrrrrr}{ReLU crosscoder sparsity and features exclusive to one input.}{tab:ext-crosscoder-exclusive}{Activation pair & $\lambda$ & $L_0$ block-sum & $L_0$ joint & A-only (\%) & B-only (\%)}
Layers 12 and 16 & 0.001 & 2707.5 & 3044.8 & 7.27 & 20.75 \\
 & 0.003 & 619.5 & 1003.9 & 0.06 & 7.11 \\
 & 0.010 & 62.9 & 104.1 & 0.02 & 1.07 \\
Step 71{,}000 and final & 0.001 & 2433.1 & 2766.4 & 13.43 & 0.64 \\
 & 0.003 & 756.3 & 1193.1 & 0.15 & 0.08 \\
 & 0.010 & 72.6 & 125.7 & 0.10 & 0.02 \\
\end{apptable}

\clearpage
\input{checklist}

\end{document}

%% file: checklist.tex
\section*{NeurIPS Paper Checklist}

\begin{enumerate}

\item {\bf Claims}
    \item[] Question: Do the main claims made in the abstract and introduction accurately reflect the paper's contributions and scope?
    \item[] Answer: \answerYes{}
    \item[] Justification: The abstract and introduction describe the measured confounds, the bypass intervention, and the scope of the experiments.
    \item[] Guidelines:
    \begin{itemize}
        \item The answer \answerNA{} means that the abstract and introduction do not include the claims made in the paper.
        \item The abstract and/or introduction should clearly state the claims made, including the contributions made in the paper and important assumptions and limitations. A \answerNo{} or \answerNA{} answer to this question will not be perceived well by the reviewers. 
        \item The claims made should match theoretical and experimental results, and reflect how much the results can be expected to generalize to other settings. 
        \item It is fine to include aspirational goals as motivation as long as it is clear that these goals are not attained by the paper. 
    \end{itemize}

\item {\bf Limitations}
    \item[] Question: Does the paper discuss the limitations of the work performed by the authors?
    \item[] Answer: \answerYes{}
    \item[] Justification: Section~\ref{sec:interpretability_probing} and Appendix~\ref{app:theory} discuss evaluation limitations and the restricted scope of the rank-one result.
    \item[] Guidelines:
    \begin{itemize}
        \item The answer \answerNA{} means that the paper has no limitation while the answer \answerNo{} means that the paper has limitations, but those are not discussed in the paper. 
        \item The authors are encouraged to create a separate ``Limitations'' section in their paper.
        \item The paper should point out any strong assumptions and how robust the results are to violations of these assumptions (e.g., independence assumptions, noiseless settings, model well-specification, asymptotic approximations only holding locally). The authors should reflect on how these assumptions might be violated in practice and what the implications would be.
        \item The authors should reflect on the scope of the claims made, e.g., if the approach was only tested on a few datasets or with a few runs. In general, empirical results often depend on implicit assumptions, which should be articulated.
        \item The authors should reflect on the factors that influence the performance of the approach. For example, a facial recognition algorithm may perform poorly when image resolution is low or images are taken in low lighting. Or a speech-to-text system might not be used reliably to provide closed captions for online lectures because it fails to handle technical jargon.
        \item The authors should discuss the computational efficiency of the proposed algorithms and how they scale with dataset size.
        \item If applicable, the authors should discuss possible limitations of their approach to address problems of privacy and fairness.
        \item While the authors might fear that complete honesty about limitations might be used by reviewers as grounds for rejection, a worse outcome might be that reviewers discover limitations that aren't acknowledged in the paper. The authors should use their best judgment and recognize that individual actions in favor of transparency play an important role in developing norms that preserve the integrity of the community. Reviewers will be specifically instructed to not penalize honesty concerning limitations.
    \end{itemize}

\item {\bf Theory assumptions and proofs}
    \item[] Question: For each theoretical result, does the paper provide the full set of assumptions and a complete (and correct) proof?
    \item[] Answer: \answerYes{}
    \item[] Justification: Appendix~\ref{app:theory} states the assumptions and provides proofs for the isolated rank-one setting.
    \item[] Guidelines:
    \begin{itemize}
        \item The answer \answerNA{} means that the paper does not include theoretical results. 
        \item All the theorems, formulas, and proofs in the paper should be numbered and cross-referenced.
        \item All assumptions should be clearly stated or referenced in the statement of any theorems.
        \item The proofs can either appear in the main paper or the supplemental material, but if they appear in the supplemental material, the authors are encouraged to provide a short proof sketch to provide intuition. 
        \item Inversely, any informal proof provided in the core of the paper should be complemented by formal proofs provided in appendix or supplemental material.
        \item Theorems and Lemmas that the proof relies upon should be properly referenced. 
    \end{itemize}

    \item {\bf Experimental result reproducibility}
    \item[] Question: Does the paper fully disclose all the information needed to reproduce the main experimental results of the paper to the extent that it affects the main claims and/or conclusions of the paper (regardless of whether the code and data are provided or not)?
    \item[] Answer: \answerYes{}
    \item[] Justification: Section~\ref{sec:setup} describes the models and data. The appendices give the SAE training and evaluation procedures.
    \item[] Guidelines:
    \begin{itemize}
        \item The answer \answerNA{} means that the paper does not include experiments.
        \item If the paper includes experiments, a \answerNo{} answer to this question will not be perceived well by the reviewers: Making the paper reproducible is important, regardless of whether the code and data are provided or not.
        \item If the contribution is a dataset and\slash or model, the authors should describe the steps taken to make their results reproducible or verifiable. 
        \item Depending on the contribution, reproducibility can be accomplished in various ways. For example, if the contribution is a novel architecture, describing the architecture fully might suffice, or if the contribution is a specific model and empirical evaluation, it may be necessary to either make it possible for others to replicate the model with the same dataset, or provide access to the model. In general. releasing code and data is often one good way to accomplish this, but reproducibility can also be provided via detailed instructions for how to replicate the results, access to a hosted model (e.g., in the case of a large language model), releasing of a model checkpoint, or other means that are appropriate to the research performed.
        \item While NeurIPS does not require releasing code, the conference does require all submissions to provide some reasonable avenue for reproducibility, which may depend on the nature of the contribution. For example
        \begin{enumerate}
            \item If the contribution is primarily a new algorithm, the paper should make it clear how to reproduce that algorithm.
            \item If the contribution is primarily a new model architecture, the paper should describe the architecture clearly and fully.
            \item If the contribution is a new model (e.g., a large language model), then there should either be a way to access this model for reproducing the results or a way to reproduce the model (e.g., with an open-source dataset or instructions for how to construct the dataset).
            \item We recognize that reproducibility may be tricky in some cases, in which case authors are welcome to describe the particular way they provide for reproducibility. In the case of closed-source models, it may be that access to the model is limited in some way (e.g., to registered users), but it should be possible for other researchers to have some path to reproducing or verifying the results.
        \end{enumerate}
    \end{itemize}

\item {\bf Open access to data and code}
    \item[] Question: Does the paper provide open access to the data and code, with sufficient instructions to faithfully reproduce the main experimental results, as described in supplemental material?
    \item[] Answer: \answerYes{}
    \item[] Justification: Code and reproduction instructions are released on GitHub, and SAE checkpoints and preprocessing assets on Hugging Face. The experiments use publicly available datasets.
    \item[] Guidelines:
    \begin{itemize}
        \item The answer \answerNA{} means that paper does not include experiments requiring code.
        \item Please see the NeurIPS code and data submission guidelines (\url{https://neurips.cc/public/guides/CodeSubmissionPolicy}) for more details.
        \item While we encourage the release of code and data, we understand that this might not be possible, so \answerNo{} is an acceptable answer. Papers cannot be rejected simply for not including code, unless this is central to the contribution (e.g., for a new open-source benchmark).
        \item The instructions should contain the exact command and environment needed to run to reproduce the results. See the NeurIPS code and data submission guidelines (\url{https://neurips.cc/public/guides/CodeSubmissionPolicy}) for more details.
        \item The authors should provide instructions on data access and preparation, including how to access the raw data, preprocessed data, intermediate data, and generated data, etc.
        \item The authors should provide scripts to reproduce all experimental results for the new proposed method and baselines. If only a subset of experiments are reproducible, they should state which ones are omitted from the script and why.
        \item At submission time, to preserve anonymity, the authors should release anonymized versions (if applicable).
        \item Providing as much information as possible in supplemental material (appended to the paper) is recommended, but including URLs to data and code is permitted.
    \end{itemize}

\item {\bf Experimental setting/details}
    \item[] Question: Does the paper specify all the training and test details (e.g., data splits, hyperparameters, how they were chosen, type of optimizer) necessary to understand the results?
    \item[] Answer: \answerYes{}
    \item[] Justification: Appendix~\ref{app:details} gives the default settings and data splits; we describe any changes per experiment.
    \item[] Guidelines:
    \begin{itemize}
        \item The answer \answerNA{} means that the paper does not include experiments.
        \item The experimental setting should be presented in the core of the paper to a level of detail that is necessary to appreciate the results and make sense of them.
        \item The full details can be provided either with the code, in appendix, or as supplemental material.
    \end{itemize}

\item {\bf Experiment statistical significance}
    \item[] Question: Does the paper report error bars suitably and correctly defined or other appropriate information about the statistical significance of the experiments?
    \item[] Answer: \answerYes{}
    \item[] Justification: We give bootstrap intervals and results across seeds; the appendices explain how we calculate the intervals and compare seeds.
    \item[] Guidelines:
    \begin{itemize}
        \item The answer \answerNA{} means that the paper does not include experiments.
        \item The authors should answer \answerYes{} if the results are accompanied by error bars, confidence intervals, or statistical significance tests, at least for the experiments that support the main claims of the paper.
        \item The factors of variability that the error bars are capturing should be clearly stated (for example, train/test split, initialization, random drawing of some parameter, or overall run with given experimental conditions).
        \item The method for calculating the error bars should be explained (closed form formula, call to a library function, bootstrap, etc.)
        \item The assumptions made should be given (e.g., Normally distributed errors).
        \item It should be clear whether the error bar is the standard deviation or the standard error of the mean.
        \item It is OK to report 1-sigma error bars, but one should state it. The authors should preferably report a 2-sigma error bar than state that they have a 96\% CI, if the hypothesis of Normality of errors is not verified.
        \item For asymmetric distributions, the authors should be careful not to show in tables or figures symmetric error bars that would yield results that are out of range (e.g., negative error rates).
        \item If error bars are reported in tables or plots, the authors should explain in the text how they were calculated and reference the corresponding figures or tables in the text.
    \end{itemize}

\item {\bf Experiments compute resources}
    \item[] Question: For each experiment, does the paper provide sufficient information on the computer resources (type of compute workers, memory, time of execution) needed to reproduce the experiments?
    \item[] Answer: \answerNo{}
    \item[] Justification: Appendix~\ref{app:compute} lists the hardware and some runtimes.
    \item[] Guidelines:
    \begin{itemize}
        \item The answer \answerNA{} means that the paper does not include experiments.
        \item The paper should indicate the type of compute workers CPU or GPU, internal cluster, or cloud provider, including relevant memory and storage.
        \item The paper should provide the amount of compute required for each of the individual experimental runs as well as estimate the total compute. 
        \item The paper should disclose whether the full research project required more compute than the experiments reported in the paper (e.g., preliminary or failed experiments that didn't make it into the paper). 
    \end{itemize}
    
\item {\bf Code of ethics}
    \item[] Question: Does the research conducted in the paper conform, in every respect, with the NeurIPS Code of Ethics \url{https://neurips.cc/public/EthicsGuidelines}?
    \item[] Answer: \answerYes{}
    \item[] Justification: The research follows the NeurIPS Code of Ethics and involves no human-subject experiments.
    \item[] Guidelines:
    \begin{itemize}
        \item The answer \answerNA{} means that the authors have not reviewed the NeurIPS Code of Ethics.
        \item If the authors answer \answerNo, they should explain the special circumstances that require a deviation from the Code of Ethics.
        \item The authors should make sure to preserve anonymity (e.g., if there is a special consideration due to laws or regulations in their jurisdiction).
    \end{itemize}

\item {\bf Broader impacts}
    \item[] Question: Does the paper discuss both potential positive societal impacts and negative societal impacts of the work performed?
    \item[] Answer: \answerNo{}
    \item[] Justification: The paper does not explicitly discuss potential positive and negative societal impacts.
    \item[] Guidelines:
    \begin{itemize}
        \item The answer \answerNA{} means that there is no societal impact of the work performed.
        \item If the authors answer \answerNA{} or \answerNo, they should explain why their work has no societal impact or why the paper does not address societal impact.
        \item Examples of negative societal impacts include potential malicious or unintended uses (e.g., disinformation, generating fake profiles, surveillance), fairness considerations (e.g., deployment of technologies that could make decisions that unfairly impact specific groups), privacy considerations, and security considerations.
        \item The conference expects that many papers will be foundational research and not tied to particular applications, let alone deployments. However, if there is a direct path to any negative applications, the authors should point it out. For example, it is legitimate to point out that an improvement in the quality of generative models could be used to generate Deepfakes for disinformation. On the other hand, it is not needed to point out that a generic algorithm for optimizing neural networks could enable people to train models that generate Deepfakes faster.
        \item The authors should consider possible harms that could arise when the technology is being used as intended and functioning correctly, harms that could arise when the technology is being used as intended but gives incorrect results, and harms following from (intentional or unintentional) misuse of the technology.
        \item If there are negative societal impacts, the authors could also discuss possible mitigation strategies (e.g., gated release of models, providing defenses in addition to attacks, mechanisms for monitoring misuse, mechanisms to monitor how a system learns from feedback over time, improving the efficiency and accessibility of ML).
    \end{itemize}
    
\item {\bf Safeguards}
    \item[] Question: Does the paper describe safeguards that have been put in place for responsible release of data or models that have a high risk for misuse (e.g., pre-trained language models, image generators, or scraped datasets)?
    \item[] Answer: \answerNA{}
    \item[] Justification: This work does not introduce any such high-risk models or data.
    \item[] Guidelines:
    \begin{itemize}
        \item The answer \answerNA{} means that the paper poses no such risks.
        \item Released models that have a high risk for misuse or dual-use should be released with necessary safeguards to allow for controlled use of the model, for example by requiring that users adhere to usage guidelines or restrictions to access the model or implementing safety filters. 
        \item Datasets that have been scraped from the Internet could pose safety risks. The authors should describe how they avoided releasing unsafe images.
        \item We recognize that providing effective safeguards is challenging, and many papers do not require this, but we encourage authors to take this into account and make a best faith effort.
    \end{itemize}

\item {\bf Licenses for existing assets}
    \item[] Question: Are the creators or original owners of assets (e.g., code, data, models), used in the paper, properly credited and are the license and terms of use explicitly mentioned and properly respected?
    \item[] Answer: \answerNo{}
    \item[] Justification: The paper does not yet explicitly list licenses and terms for all datasets, models, and code used.
    \item[] Guidelines:
    \begin{itemize}
        \item The answer \answerNA{} means that the paper does not use existing assets.
        \item The authors should cite the original paper that produced the code package or dataset.
        \item The authors should state which version of the asset is used and, if possible, include a URL.
        \item The name of the license (e.g., CC-BY 4.0) should be included for each asset.
        \item For scraped data from a particular source (e.g., website), the copyright and terms of service of that source should be provided.
        \item If assets are released, the license, copyright information, and terms of use in the package should be provided. For popular datasets, \url{paperswithcode.com/datasets} has curated licenses for some datasets. Their licensing guide can help determine the license of a dataset.
        \item For existing datasets that are re-packaged, both the original license and the license of the derived asset (if it has changed) should be provided.
        \item If this information is not available online, the authors are encouraged to reach out to the asset's creators.
    \end{itemize}

\item {\bf New assets}
    \item[] Question: Are new assets introduced in the paper well documented and is the documentation provided alongside the assets?
    \item[] Answer: \answerYes{}
    \item[] Justification: The code and SAE checkpoints are accompanied by training configurations, usage instructions, licenses, and documentation of known limitations.
    \item[] Guidelines:
    \begin{itemize}
        \item The answer \answerNA{} means that the paper does not release new assets.
        \item Researchers should communicate the details of the dataset\slash code\slash model as part of their submissions via structured templates. This includes details about training, license, limitations, etc. 
        \item The paper should discuss whether and how consent was obtained from people whose asset is used.
        \item At submission time, remember to anonymize your assets (if applicable). You can either create an anonymized URL or include an anonymized zip file.
    \end{itemize}

\item {\bf Crowdsourcing and research with human subjects}
    \item[] Question: For crowdsourcing experiments and research with human subjects, does the paper include the full text of instructions given to participants and screenshots, if applicable, as well as details about compensation (if any)? 
    \item[] Answer: \answerNA{}
    \item[] Justification: The research involves no crowdsourcing or human participants.
    \item[] Guidelines:
    \begin{itemize}
        \item The answer \answerNA{} means that the paper does not involve crowdsourcing nor research with human subjects.
        \item Including this information in the supplemental material is fine, but if the main contribution of the paper involves human subjects, then as much detail as possible should be included in the main paper. 
        \item According to the NeurIPS Code of Ethics, workers involved in data collection, curation, or other labor should be paid at least the minimum wage in the country of the data collector. 
    \end{itemize}

\item {\bf Institutional review board (IRB) approvals or equivalent for research with human subjects}
    \item[] Question: Does the paper describe potential risks incurred by study participants, whether such risks were disclosed to the subjects, and whether Institutional Review Board (IRB) approvals (or an equivalent approval/review based on the requirements of your country or institution) were obtained?
    \item[] Answer: \answerNA{}
    \item[] Justification: The research involves no human participants.
    \item[] Guidelines:
    \begin{itemize}
        \item The answer \answerNA{} means that the paper does not involve crowdsourcing nor research with human subjects.
        \item Depending on the country in which research is conducted, IRB approval (or equivalent) may be required for any human subjects research. If you obtained IRB approval, you should clearly state this in the paper. 
        \item We recognize that the procedures for this may vary significantly between institutions and locations, and we expect authors to adhere to the NeurIPS Code of Ethics and the guidelines for their institution. 
        \item For initial submissions, do not include any information that would break anonymity (if applicable), such as the institution conducting the review.
    \end{itemize}

\item {\bf Declaration of LLM usage}
    \item[] Question: Does the paper describe the usage of LLMs if it is an important, original, or non-standard component of the core methods in this research? Note that if the LLM is used only for writing, editing, or formatting purposes and does \emph{not} impact the core methodology, scientific rigor, or originality of the research, declaration is not required.
    \item[] Answer: \answerYes{}
    \item[] Justification: Section~\ref{sec:interpretability_probing} and Appendix~\ref{app:rx_autointerp} describe the LLM explainers, evaluators, and scoring procedure.
    \item[] Guidelines:
    \begin{itemize}
        \item The answer \answerNA{} means that the core method development in this research does not involve LLMs as any important, original, or non-standard components.
        \item Please refer to our LLM policy in the NeurIPS handbook for what should or should not be described.
    \end{itemize}

\end{enumerate}